%% file: aaai2026.tex
\documentclass[letterpaper]{article} % DO NOT CHANGE THIS
\usepackage{aaai2026}
\usepackage{times}  % DO NOT CHANGE THIS
\usepackage{helvet}  % DO NOT CHANGE THIS
\usepackage{courier}  % DO NOT CHANGE THIS
\usepackage[hyphens]{url}  % DO NOT CHANGE THIS
\usepackage{graphicx} % DO NOT CHANGE THIS
\usepackage{natbib}  % DO NOT CHANGE THIS AND DO NOT ADD ANY OPTIONS TO IT
\usepackage{caption} % DO NOT CHANGE THIS AND DO NOT ADD ANY OPTIONS TO IT
\usepackage{algorithm}
\usepackage{algorithmic}
\usepackage{booktabs} 
\usepackage{amsmath}

\usepackage{newfloat}
\usepackage{listings}
\DeclareCaptionStyle{ruled}{labelfont=normalfont,labelsep=colon,strut=off} % DO NOT CHANGE THIS
\floatstyle{ruled}
\newfloat{listing}{tb}{lst}{}
\floatname{listing}{Listing}
\title{\textsc{JudgeBoard}: Benchmarking and Enhancing Small Language Models for Reasoning Evaluation}

\author{
Zhenyu Bi\textsuperscript{1}, Gaurav Srivastava\textsuperscript{1}, Yang Li\textsuperscript{2}, Meng Lu\textsuperscript{1}, 
Swastik Roy\textsuperscript{3},\\ Morteza Ziyadi\textsuperscript{3}, Xuan Wang\textsuperscript{1}
}

\affiliations {
    \textsuperscript{\rm 1}Virginia Tech \\
    \textsuperscript{\rm 2}College of William and Mary \\
    \textsuperscript{\rm 3}Amazon AGI\\
    \textsuperscript{\rm 1} \{zhenyub,gks,menglu\}@vt.edu, \textsuperscript{\rm 2} yli102@wm.edu,\textsuperscript{\rm 3} \{roswasti, mziyadi\}@amazon.com
    
}

\usepackage{bibentry}
\begin{document}

\maketitle

\begin{abstract}
While small language models (SLMs) have shown promise on various reasoning tasks, their ability to judge the correctness of answers remains unclear compared to large language models (LLMs). Prior work on LLM-as-a-judge frameworks typically relies on comparing candidate answers against ground-truth labels or other candidate answers using predefined metrics like entailment. However, this approach is inherently indirect and difficult to fully automate, offering limited support for fine-grained and scalable evaluation of reasoning outputs. In this work, we propose JudgeBoard, a novel evaluation pipeline that directly queries models to assess the correctness of candidate answers without requiring extra answer comparisons. We focus on two core reasoning domains: mathematical reasoning and science/commonsense reasoning, and construct task-specific evaluation leaderboards using both accuracy-based ranking and an Elo-based rating system across five benchmark datasets, enabling consistent model comparison as judges rather than comparators. To improve judgment performance in lightweight models, we propose MAJ (Multi-Agent Judging), a novel multi-agent evaluation framework that leverages multiple interacting SLMs with distinct reasoning profiles to approximate LLM-level judgment accuracy through collaborative deliberation. Experimental results reveal a significant performance gap between SLMs and LLMs in isolated judging tasks. However, our MAJ framework substantially improves the reliability and consistency of SLMs. On the MATH dataset, MAJ using smaller-sized models as backbones performs comparatively well or even better than their larger-sized counterparts. Our findings highlight that multi-agent SLM systems can potentially match or exceed LLM performance in judgment tasks, with implications for scalable and efficient assessment.
\end{abstract}


\section{Introduction}
Large language models (LLMs) have demonstrated remarkable capabilities in a wide range of natural language processing tasks, including reasoning, question answering, and evaluation \cite{achiam2023gpt,dubey2024llama,liu2024deepseek,yang2025qwen3}. A growing body of work has explored the use of LLMs not only as generators of content but also as evaluators, where models are tasked with assessing the relative quality of candidate outputs \cite{zheng2023judging,tan2024judgebench}. These approaches have been applied in various reasoning domains, often relying on indirect supervision via entailment metrics or comparisons to gold-standard labels or the other candidate outputs (llm-as-a-judge). While effective in some settings, these methods are inherently limited: they rely on comparison-based depend on predefined metrics that may not capture nuanced reasoning errors, and they often require human-curated labels, which hinders scalability and generalization.

In contrast, small language models (SLMs) offer a more efficient and accessible alternative to LLMs \cite{yang2025qwen3,abdin2024phi,team2024gemma, srivastava-etal-2025-thinkslm, srivastava-etal-2025-debate, srivastava2025llmsoverthinkbasicmath}, particularly in resource-constrained environments. Recent studies have shown that SLMs can perform competitively on various reasoning tasks when equipped with appropriate prompting or fine-tuning strategies \cite{wei2022chain}. However, their ability to judge the correctness of answers rather than generate them remains underexplored. This gap is critical, as scalable and reliable evaluation is essential for deploying language models in real-world applications.

\begin{figure}[t]
\centering
\includegraphics[width=1\columnwidth]{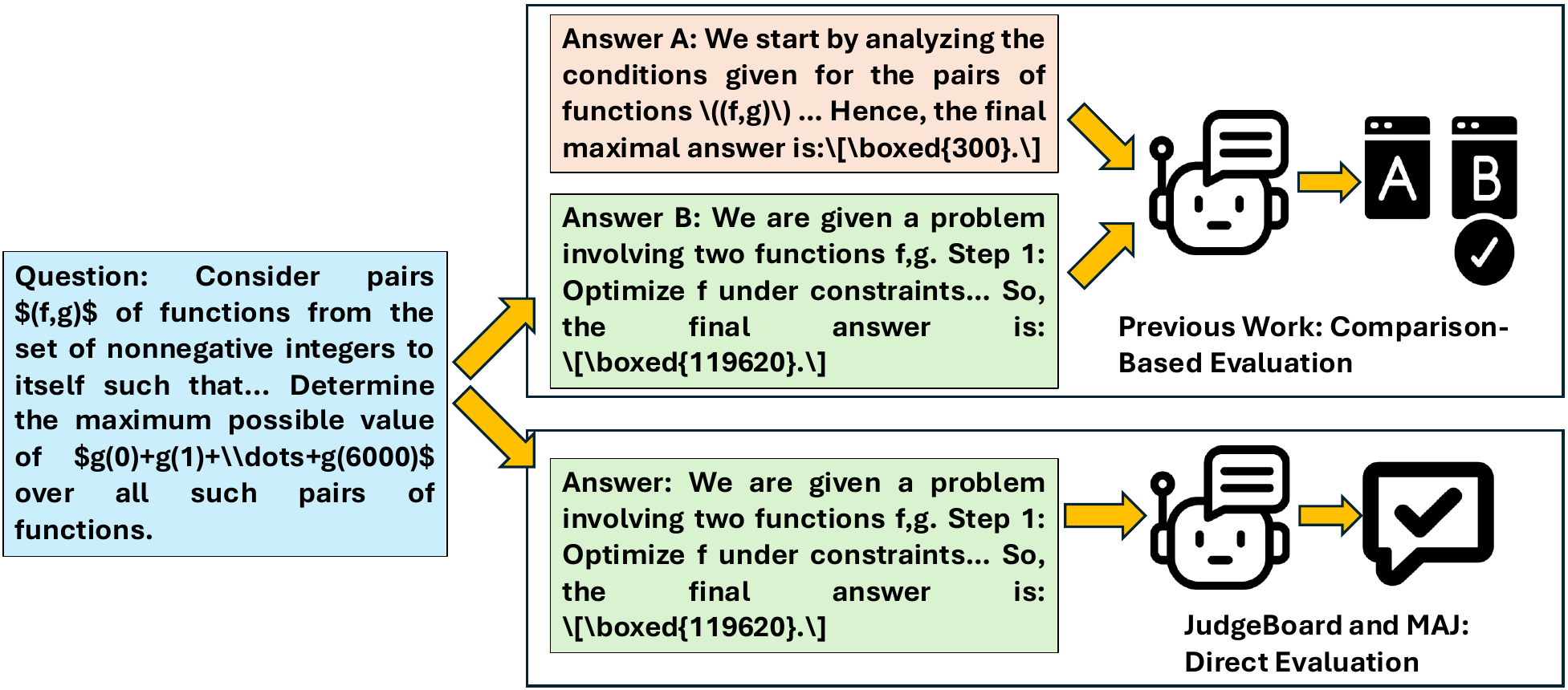} 
\caption{Comparison of JudgeBoard and MAJ with previous works. Unlike previous works that usually follow a comparison-based evaluation pipeline, JudgeBoard and MAJ focus on direct evaluation of the factual correctness of the reasoning questions.}
\label{Fig:compare}
\end{figure}

In this work, we propose JudgeBoard, a novel evaluation pipeline that directly queries models, particularly SLMs, to assess the correctness of candidate answers without requiring extra answer comparisons. Figure \ref{Fig:compare} demonstrates the difference between our proposed pipeline and prior works. We focus on two core domains: mathematical reasoning and science/commonsense reasoning. To support this setup, we construct task-specific evaluation leaderboards across five benchmark datasets, using both accuracy-based rankings and an Elo-style rating system to enable consistent and fine-grained comparison of models in their role as judges. The results highlight the need for using multiple evaluation metrics when doing evaluation, provide insights on the judging abilities of different SLMs, and demonstrate the significant performance gap between the best-performing Large Language Models and the best-performing Small Language Models (SLMs).

To address the limitations of individual SLMs in judgment tasks, we introduce MAJ (Multi-Agent Judging), a novel Elo-based framework that leverages multiple interacting SLMs to approximate the judgment performance of LLMs. Inspired by recent advances in multi-agent collaboration and self-consistency techniques \cite{zhugegptswarm, guo2024large, wu2024autogen}, MAJ enables lightweight models to collectively reason about answer correctness, improving both reliability and robustness. Our experiments reveal a substantial performance gap between isolated SLMs and LLMs in judgment tasks, but also show that MAJ significantly narrows this gap. On the MATH dataset \cite{lightman2023lets}, our MAJ framework using Qwen3-14B model as the backbone outperform the best-performing LLM by an average of 2\% in judging accuracy across all categories. Our findings suggest that with the right collaborative framework, SLMs can rival or even surpass LLMs in evaluative reasoning; this finding paves the way for more efficient and democratized model assessment.

\section{Related Works}
\paragraph{Language Models as Judges}
LLMs-as-a-judge frameworks have emerged as an alternative to human annotators and traditional metrics. \citeauthor{tan2024judgebench} evaluated 11 LLMs across 20 NLP tasks, finding variable reliability depending on task type and data source. \citeauthor{tang2024large} surveyed the paradigm across functionality, methodology, applications, meta-evaluation, and limitations, highlighting interpretability benefits while cautioning against biases in proprietary models. Practical implementations include MT-Bench and Chatbot Arena \cite{zheng2023judging}, which demonstrate that LLMs can approximate human evaluators with proper calibration.
\paragraph{LLM Reasoning Prompting}
Prompting techniques have enhanced LLM reasoning capabilities. Chain-of-Thought (CoT) \cite{wei2022chain} breaks reasoning into sequential steps, while self-consistency \cite{wang2022self} aggregates multiple reasoning paths. Tree-of-Thought (ToT) \cite{yao2023tree} maintains structured intermediate steps, and Graph-of-Thought \cite{besta2024graph} uses directed graphs to revisit and combine reasoning paths. Additional X-of-Thought variants address specific tasks \cite{pot,aot,myself}. Recent reasoning benchmarks \cite{srivastava2025beyondbenchbenchmarkfreeevaluationreasoning, srivastava2025llmsoverthinkbasicmath} also evaluate and compare the effect of different reasoning prompting methods on LLMs.

\paragraph{Multi-Agent Reasoning}
Multi-agent LLM frameworks have shown improved reasoning over single-agent systems \cite{wu2024autogen, chen2023autoagents, menglu}. Research on multi-agent debate dynamics \cite{Wang2023ASO, Wang2024RethinkingTB, Pezeshkpour2024ReasoningCI} reveals that most interaction protocols are manually defined \cite{Wu2023AutoGenEN, Chan2023ChatEvalTB} or follow simple formats like majority voting and summarization \cite{Chen2023ReConcileRC, liang2024encouragingdivergentthinkinglarge, Chan2023ChatEvalTB}. GPTSwarm and OptAgent \cite{zhugegptswarm,bi2025optagentoptimizingmultiagentllm} models multi-agent systems as graph networks, enabling algorithmic optimization of interaction patterns and agent-level prompts.

\paragraph{Small Language Models}
Small Language Models (SLMs) offer efficient alternatives for reasoning and evaluation with reduced computational costs. Recent instruction-tuned SLMs with fewer than 3 billion parameters achieve strong reasoning capabilities through high-quality training data \cite{gunasekar2023textbooks,abdin2024phi, bai2023qwen, team2024gemma}. Recent work \cite{srivastava-etal-2025-thinkslm, srivastava-etal-2025-debate} explored methods to enhance SLMs' reasoning abilities through targeted training and architectural improvements. \citeauthor{tang2024rethinking} found that SLMs can effectively judge reasoning quality with appropriate prompting, while \citeauthor{liu2024makes} identified key factors enabling effective reasoning, including training data quality and model architecture. Efficient reasoning frameworks \cite{shao2024deepseekmath} demonstrate that training on mathematical and logical reasoning data enables SLMs to compete with larger models.

\begin{figure*}[t]
     \centering
    \includegraphics[width=0.8\linewidth]{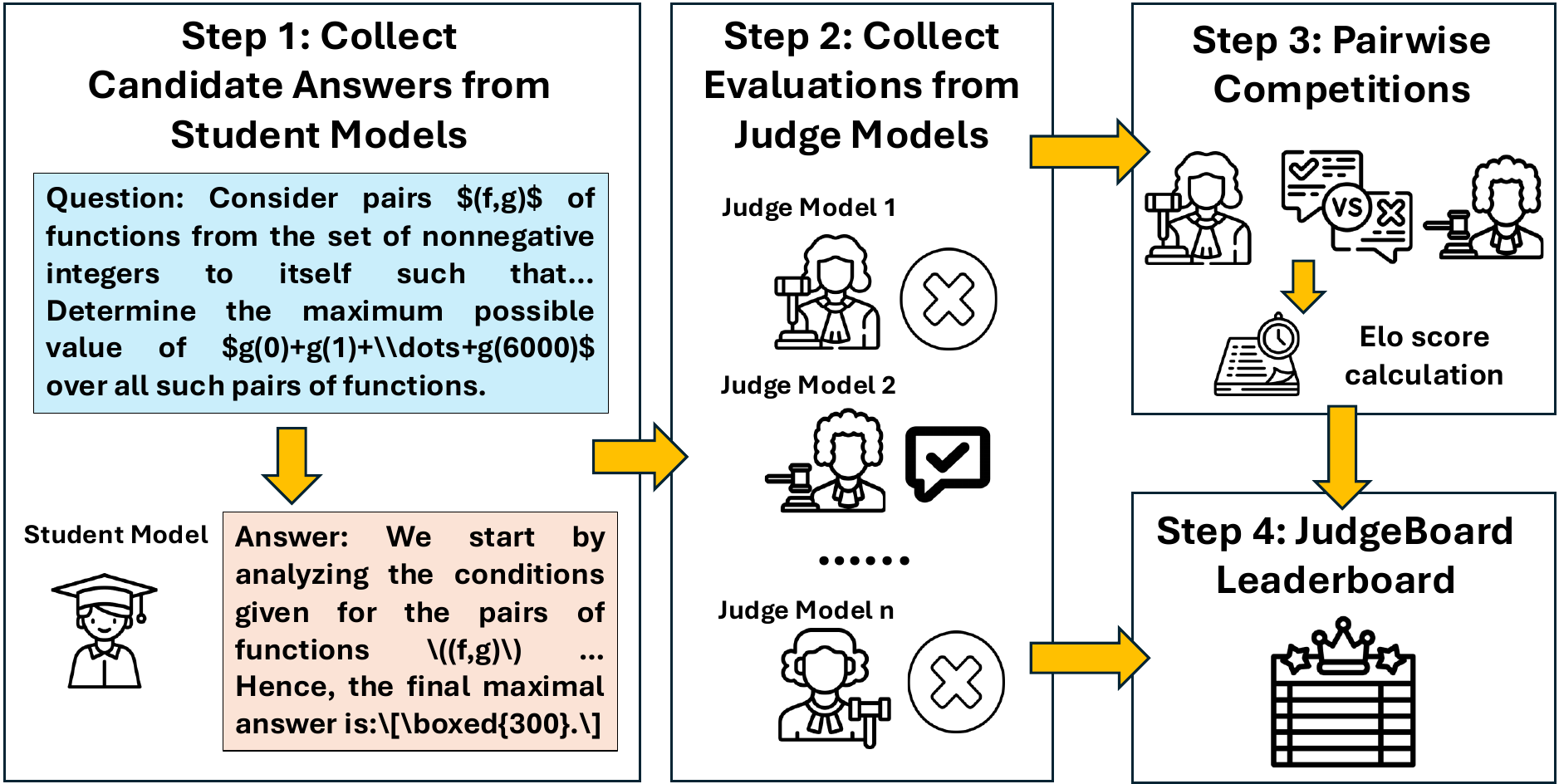}
     \caption{Overview of the JudgeBoard Pipeline}\label{Fig:overview}
\hfill
\end{figure*}
\section{JudgeBoard}
\subsection{Pipeline Overview}
We propose a novel evaluation pipeline, \textbf{JudgeBoard}, that evaluates the models' judging ability by querying models to assess the correctness of candidate answers without requiring extra answer comparisons. This approach treats models as direct evaluators, enabling scalable and flexible assessment across diverse reasoning tasks. While our judging protocol is direct, meta-evaluation of judge quality necessarily requires gold-standard labels for ground truth. The overall pipeline of JudgeBoard is demonstrated in Figure \ref{Fig:overview}. JudgeBoard consists of four stages:
\begin{itemize}
\item \textbf{Candidate Answer Collection}: Student model is provided with a set of reasoning questions and is prompted to provide answers.
\item \textbf{Judgment Collection}: A set of judge models independently evaluate the correctness of each candidate answer.
\item \textbf{Pairwise Competitions}: For the same question-answer pair, evaluations from different judges are compared in a pairwise fashion. We then calculate the Elo score based on the competition results.
\item \textbf{Leaderboard Construction}: We compute two types of rankings: Accuracy-based ranking and Elo-style-based ranking, to assess and compare model performance across tasks and subcategories.
\end{itemize}

\subsection{Model-as-Judge Protocol}
To operationalize the Model-as-Judge pipeline, we design a structured prompting protocol that ensures consistency and objectivity in model evaluations. Each judge model is presented with the original question, the candidate answer generated by the student model, and a prompt instructing the judge to determine whether the answer is correct or incorrect and output its reasoning. Prompts are carefully engineered to minimize ambiguity and bias, and to encourage factual, step-by-step reasoning. 

Additionally, we ensure similar but different reasoning by assigning the agents with the same baseline reasoning prompt but different agent profiles in system prompts (see Appendix A for detailed prompts). The agent profiles were manually crafted to reflect common reasoning strategies found in human problem-solving, such as deductive reasoning, logical reasoning, and robust reasoning. 

\subsection{Pairwise Evaluation and Elo Rating Calculation}
\label{elo}
After collecting judgments, we conduct pairwise comparisons between candidate answers. For each question, judgments of the answers are evaluated against each other. We determine outcomes based on agreement with gold-standard labels: a judge wins if its judgment matches the gold label while its comparator does not. To quantitatively assess the relative performance of judge models, we adopt an Elo-style rating system inspired by the ChatbotArena.

The Elo system provides additional value beyond simple accuracy metrics through three key mechanisms. First, it accounts for question difficulty by awarding judges more credit for correct judgments on questions where other judges failed, effectively weighting performance by task complexity. Second, it measures consistency across diverse question types, rewarding models that reliably agree with gold labels regardless of domain or difficulty rather than those that perform well only on specific subsets. Finally, the dynamic rating system better distinguishes models with similar raw accuracy by capturing the relative strength demonstrated through head-to-head comparisons, revealing nuanced performance differences that aggregate accuracy scores may obscure. This enables a more granular evaluation of model capabilities across reasoning tasks.

\paragraph{Rating Update Mechanism} Let $R_i$ and $R_j$ denote the current Elo ratings of models i and j, respectively. The expected score for model i is computed using the standard logistic function: $E_i = \frac{1}{1 + 10^{(R_j - R_i)/400}}$. Following the outcome of the match, the rating of model i is updated by $R_i' = R_i + K(S_i - E_i)$, where $R_i$ is the rating of model i, $S_i \in \{1, 0.5, 0\}$ represents the actual match outcome (win, draw, loss), and $K=10$ is a constant controlling the magnitude of rating updates. All models are initialized with a uniform baseline rating. To ensure robustness, each model participates in a large number of comparisons across diverse questions and judge profiles. The final Elo score reflects the model’s overall performance relative to its peers.

\subsection{Leaderboard Construction}
We construct task-specific leaderboards using two complementary metrics:
\paragraph{Accuracy-Based Ranking} Measures agreement with the gold-label given by the original dataset across three dimensions: Overall Accuracy (correctness across all questions), Student Wrong (SW) Accuracy (ability to identify incorrect answers), and Student Right (SR) Accuracy (ability to validate correct answers). These metrics reveal not only judging performance but also potential sycophacy toward agreement.
\paragraph{Elo-Style Rating System} Models are evaluated in pairwise comparisons, and ratings are updated based on relative performance. This enables fine-grained differentiation and comparison even in ambiguous cases.

\begin{table*}[ht]
\centering
\setlength{\tabcolsep}{1mm}
\begin{tabular}{@{}l*{3}{ccc}@{}}
\toprule
& \multicolumn{3}{c}{\textbf{Algebra}} & \multicolumn{3}{c}{\textbf{Number Theory}} & \multicolumn{3}{c}{\textbf{Counting and Probability}} \\
\cmidrule(lr){2-4} \cmidrule(lr){5-7} \cmidrule(lr){8-10}
\textbf{Model} & \textbf{Overall} & \textbf{SW} & \textbf{Elo} & \textbf{Overall} & \textbf{SW} & \textbf{Elo} & \textbf{Overall} & \textbf{SW} & \textbf{Elo} \\
& \textbf{Accuracy} & \textbf{Accuracy} & & \textbf{Accuracy} & \textbf{Accuracy} & & \textbf{Accuracy} & \textbf{Accuracy} & \\
\midrule
\multicolumn{10}{c}{\textit{Large Language Models ($>$14B)}} \\
\midrule
Qwen3\_30B\_A3B & 0.7789 & 0.570 & 1070.8 & 0.9077 & 0.825 & 1110.3 & 0.8889 & 0.793 & 1089.3 \\
Qwen3\_32B & 0.7632 & 0.516 & 1068.8 & 0.8846 & 0.778 & 1100.6 & 0.8888 & 0.782 & 1091.4 \\
%Claude 3.7 & 0.7421 & 0.495 & 1056.7 & 0.8692 & 0.730 & 1094.2 & 0.8667 & 0.793 & 1080.9 \\
Gemma3-27b & 0.6947 & 0.398 & 1038.8 & 0.8077 & 0.603 & 1069.2 & 0.7944 & 0.621 & 1054.4 \\
Qwen2.5-72b & 0.6368 & 0.269 & 1017.2 & 0.6846 & 0.365 & 1020.9 & 0.7167 & 0.471 & 1026.4 \\
Llama3-70b & 0.6053 & 0.215 & 1005.5 & 0.6462 & 0.270 & 1006.0 & 0.7056 & 0.414 & 1022.5 \\
Llama4-17B & 0.5895 & 0.183 & 999.6 & 0.7000 & 0.381 & 1026.9 & 0.6778 & 0.356 & 1012.6 \\
Mixtral-8x22B & 0.5579 & 0.097 & 987.9 & 0.5462 & 0.095 & 973.2 & 0.5667 & 0.126 & 977.0 \\
Mixtral-8x7B & 0.5316 & 0.054 & 978.2 & 0.5692 & 0.111 & 979.2 & 0.5389 & 0.069 & 967.0 \\
\midrule
\multicolumn{10}{c}{\textit{Small Language Models ($<$14B)}} \\
\midrule
Gemma3\_12b & 0.6737 & 0.366 & 1030.9 & 0.7923 & 0.571 & 1063.1 & 0.7611 & 0.575 & 1042.4 \\
Qwen3\_14B & 0.6684 & 0.344 & 1030.9 & 0.6923 & 0.365 & 1020.9 & 0.7111 & 0.506 & 1024.5 \\
Phi4 & 0.6737 & 0.344 & 1030.9 & 0.7077 & 0.413 & 1032.9 & 0.7056 & 0.425 & 1022.5 \\
Gemma3\_4b & 0.6526 & 0.323 & 1021.1 & 0.7461 & 0.556 & 1038.8 & 0.7444 & 0.540 & 1036.4 \\
Qwen3\_4B & 0.6421 & 0.280 & 1019.2 & 0.7384 & 0.492 & 1041.9 & 0.7444 & 0.540 & 1036.4 \\
Qwen3\_8B & 0.5737 & 0.129 & 976.2 & 0.6308 & 0.238 & 958.2 & 0.6667 & 0.356 & 988.9 \\
Llama3.1\_8B & 0.5211 & 0.054 & 970.3 & 0.5385 & 0.093 & 961.2 & 0.5667 & 0.138 & 967.0 \\
Mistral\_7B & 0.5211 & 0.032 & 978.2 & 0.5308 & 0.048 & 964.2 & 0.5555 & 0.081 & 971.0 \\
Llama3.2\_3B & 0.5158 & 0.032 & 964.4 & 0.5308 & 0.032 & 958.2 & 0.5389 & 0.069 & 963.0 \\
Qwen3\_1.7B & 0.5789 & 0.151 & 995.7 & 0.5615 & 0.111 & 958.2 & 0.5944 & 0.161 & 965.0 \\
Qwen3\_0.6B & 0.5211 & 0.022 & 974.2 & 0.5462 & 0.064 & 967.2 & 0.5555 & 0.092 & 969.0 \\
Qwen3\_4B\_Reasoning & 0.6474 & 0.323 & 1023.1 & 0.7000 & 0.413 & 1026.9 & 0.7000 & 0.437 & 1020.5 \\
Qwen3\_14B\_Reasoning & 0.6474 & 0.301 & 1019.2 & 0.7077 & 0.397 & 1032.9 & 0.7222 & 0.471 & 1030.4 \\
Qwen3\_8B\_Reasoning & 0.5895 & 0.204 & 1001.6 & 0.6923 & 0.413 & 1023.9 & 0.6389 & 0.333 & 998.8 \\
\bottomrule
\end{tabular}%
\caption{JudgeBoard Leaderboard Results for Base Language Models. "Overall Accuracy" represents the overall judging accuracy of the judge models; "SW Accuracy" represents the judging accuracy of the judge models when student model have made a wrong answer; "Elo" represents the elo score of the juge model. This comprehensive evaluation compares the performance of large language models ($>$14B parameters) and small language models ($<$14B parameters) across three mathematical domains: Algebra, Number Theory, and Counting \& Probability.}
\label{tab:judgebench_base}
\end{table*}

\begin{table*}[ht]
\centering
%\resizebox{\textwidth}{!}{%
\begin{tabular}{@{}l*{3}{cc}@{}}
\toprule
& \multicolumn{2}{c}{\textbf{Algebra}} & \multicolumn{2}{c}{\textbf{Number Theory}} & \multicolumn{2}{c}{\textbf{Counting and Probability}} \\
\cmidrule(lr){2-3} \cmidrule(lr){4-5} \cmidrule(lr){6-7}
\textbf{Model} & \textbf{Overall} & \textbf{Accuracy -} & \textbf{Overall} & \textbf{Accuracy -} & \textbf{Overall} & \textbf{Accuracy -} \\
& \textbf{Accuracy} & \textbf{Student Wrong} & \textbf{Accuracy} & \textbf{Student Wrong} & \textbf{Accuracy} & \textbf{Student Wrong} \\
\midrule
\multicolumn{7}{c}{\textit{Multi-Agent Profiled Debate Systems}} \\
\midrule
Qwen3\_14B & 0.7789 & 0.591 & 0.9385 & 0.873 & 0.9056 & 0.851 \\
Qwen3\_8B & 0.7632 & 0.538 & 0.9 & 0.81 & 0.8889 & 0.751 \\
Qwen3\_4B & 0.7632 & 0.527 & 0.9 & 0.794 & 0.8444 & 0.713 \\
Gemma3\_12b & 0.7 & 0.441 & 0.7231 & 0.413 & 0.7167 & 0.437 \\
Gemma3\_4b & 0.6421 & 0.29 & 0.7231 & 0.444 & 0.7944 & 0.724 \\
Phi4 & 0.6421 & 0.29 & 0.6692 & 0.333 & 0.7278 & 0.471 \\
\bottomrule
\end{tabular}%
%}
\caption{JudgeBoard Leaderboard Results for Multi-Agent Jury Systems. This table presents the performance of collaborative multi-agent systems that employ profiled debate methodologies with majority voting mechanisms.}
\label{tab:judgebench_multiagent}
\end{table*}

\section{Multi-Agent Judging with Profiling}
To enhance the robustness and performance of Small Language Model evaluation, we introduce a multi-agent judge framework in which multiple SLMs independently assess the correctness of candidate answers. Each model, referred to as an agent, operates autonomously and contributes to a collective judgment through structured debate and interaction. This setup allows us not only to mitigate the biases of individual models but also to analyze their evaluative behavior through profiling.

\begin{table}[t]
\centering
\begin{tabular}{lccc}
\hline
\textbf{Model} & \textbf{Overall} & \textbf{SW} & \textbf{Elo} \\
& \textbf{Accuracy} & \textbf{Accuracy} & \textbf{Score} \\
\hline
\multicolumn{4}{l}{\textit{Large Language Models ($>$14B)}} \\
\hline
Gemma3\_27b & 0.880 & 0.430 & 1021.1 \\
Qwen2.5\_72b & 0.870 & 0.500 & 1038.7 \\
Llama3.3\_70b & 0.850 & 0.210 & 992.0 \\
Llama4\_17B & 0.850 & 0.320 & 997.8 \\
Mixtral\_8x22B & 0.850 & 0.210 & 992.0 \\
Mixtral\_8x7B & 0.820 & 0.110 & 983.1 \\
\hline
\multicolumn{4}{l}{\textit{Small Language Models ($<$14B)}} \\
\hline
Gemma3\_12b & 0.850 & 0.360 & 1015.3 \\
Qwen3\_14B & 0.870 & 0.540 & 1032.8 \\
Phi4 & 0.850 & 0.320 & 1009.5 \\
Gemma3\_4b & 0.810 & 0.070 & 968.7 \\
Qwen3\_4B & 0.840 & 0.320 & 1003.6 \\
Qwen3\_8B & 0.850 & 0.460 & 1027.0 \\
Llama3.1\_8B & 0.780 & 0.390 & 980.4 \\
Llama3.2\_3B & 0.650 & 0.360 & 968.7 \\
Qwen3\_1.7B & 0.810 & 0.320 & 1009.5 \\
Qwen3\_0.6B & 0.790 & 0.140 & 974.5 \\
\hline
\end{tabular}
\caption{JudgeBoard Leaderboard Results for Base Language Models on ARC-Challenge Dataset. ARC-Challenge contains grade school science questions requiring reasoning.}
\label{tab:arc_results}
\end{table}

\begin{table}[t]
\centering
\begin{tabular}{@{}lccc@{}}
\toprule
\textbf{Model} & \textbf{Algebra} & \textbf{NumTheo} & \textbf{C\&P} \\
\midrule
\multicolumn{4}{c}{\textit{Deductive Reasoning (DR) Judges}} \\
\midrule
Gemma3\_12b & 1029.0 & 1066.2 & 1038.4 \\
Gemma3\_4b & 1030.9 & 1032.9 & 1026.4 \\
Llama3.1\_8B & 970.3 & 964.2 & 967.0 \\
Llama3.2\_3B & 964.4 & 949.1 & 951.0 \\
Mistral\_7B & 974.2 & 970.2 & 969.0 \\
Phi4 & 1015.3 & 1014.9 & 1024.5 \\
Qwen3\_14B & 1013.3 & 1020.9 & 1026.4 \\
Qwen3\_4B & 1029.0 & 1020.9 & 1020.5 \\
Qwen3\_8B & 968.3 & 933.8 & 976.6 \\
\midrule
\multicolumn{4}{c}{\textit{Logical Reasoning (LR) Judges}} \\
\midrule
Gemma3\_12b & 1027.0 & 1047.9 & 1032.4 \\
Gemma3\_4b & 1011.4 & 1038.8 & 1018.5 \\
Llama3.1\_8B & 853.1 & 841.4 & 835.4 \\
Llama3.2\_3B & 966.4 & 964.2 & 963.0 \\
Mistral\_7B & 972.3 & 970.2 & 969.0 \\
Phi4 & 1023.1 & 1009.0 & 1030.4 \\
Qwen3\_14B & 1017.2 & 1012.0 & 1020.5 \\
Qwen3\_4B & 980.1 & 997.1 & 1014.6 \\
Qwen3\_8B & 942.6 & 930.7 & 949.0 \\
\midrule
\multicolumn{4}{c}{\textit{Robust Reasoning (RR) Judges}} \\
\midrule
Gemma3\_12b & 1029.0 & 1063.1 & 1032.4 \\
Gemma3\_4b & 993.8 & 1006.0 & 1024.5 \\
Llama3.1\_8B & 962.4 & 952.1 & 961.0 \\
Llama3.2\_3B & 946.6 & 943.0 & 932.7 \\
Mistral\_7B & 974.2 & 967.2 & 967.0 \\
Phi4 & 1021.1 & 1026.9 & 1026.4 \\
Qwen3\_14B & 1050.4 & 1014.9 & 1032.4 \\
Qwen3\_4B & 1013.3 & 1020.9 & 1032.4 \\
Qwen3\_8B & 950.5 & 933.8 & 963.0 \\
\bottomrule
\end{tabular}%
\caption{JudgeBoard Leaderboard of models prompted by different profiles, in terms of Elo Scores; Numtheo represent the Number Theory category, and C\&P represent the Counting and Probability category.}
\label{fig:judgebench_reasoning_sorted}
\end{table}

% \begin{table}[t]
% \centering
% \begin{tabular}{lcc}
% \toprule
% \textbf{Model} & \textbf{Overall} & \textbf{Accuracy -} \\
% & \textbf{Accuracy} & \textbf{Student Wrong} \\
% \midrule
% \multicolumn{3}{l}{\textit{Algebra}} \\
% \midrule
% Qwen3\_14B & 0.7789 & 0.591 \\
% Qwen3\_8B & 0.7632 & 0.538 \\
% Qwen3\_4B & 0.7632 & 0.527 \\
% Gemma3\_12b & 0.7 & 0.441 \\
% Gemma3\_4b & 0.6421 & 0.29 \\
% Phi4 & 0.6421 & 0.29 \\
% \midrule
% \multicolumn{3}{l}{\textit{Number Theory}} \\
% \midrule
% Qwen3\_14B & 0.9385 & 0.873 \\
% Qwen3\_8B & 0.9 & 0.81 \\
% Qwen3\_4B & 0.9 & 0.794 \\
% Gemma3\_12b & 0.7231 & 0.413 \\
% Gemma3\_4b & 0.7231 & 0.444 \\
% Phi4 & 0.6692 & 0.333 \\
% \midrule
% \multicolumn{3}{l}{\textit{Counting and Probability}} \\
% \midrule
% Qwen3\_14B & 0.9056 & 0.851 \\
% Qwen3\_8B & 0.8889 & 0.751 \\
% Qwen3\_4B & 0.8444 & 0.713 \\
% Gemma3\_12b & 0.7167 & 0.437 \\
% Gemma3\_4b & 0.7944 & 0.724 \\
% Phi4 & 0.7278 & 0.471 \\
% \bottomrule
% \end{tabular}
% \caption{JudgeBoard Leaderboard Results for Multi-Agent Jury Systems. This table presents the performance of collaborative multi-agent systems that employ profiled debate methodologies with majority voting mechanisms.}
% \label{tab:judgebench_multiagent}
% \end{table}

\subsection{Agent Configuration and Profiling}
Each agent is instantiated from a distinct language model instance, ensuring diversity in reasoning while maintaining a consistent evaluation framework. To promote varied yet comparable analytical behavior, all agents are initialized with a shared baseline reasoning prompt that outlines the general task and expectations. However, each agent is also assigned a unique system prompt, which is referred to as its profile, that guides its reasoning style and evaluative priorities (We provide the details of the prompts in Appendix A). The agent profiles were manually crafted to reflect common reasoning strategies found in human problem-solving. 

During evaluation, each agent receives the original question and a candidate answer and is asked to determine whether the answer is correct. In addition to a binary judgment (correct/incorrect), agents are prompted to provide a concise natural language explanation that justifies their decision. These explanations serve two purposes: they make the reasoning process interpretable and they form the basis for subsequent inter-agent debate.

\subsection{Interaction and Debate}
After the agents get their initial answers, we let them discuss with each other in order to find a better answer. We implement a structured multi-turn debate protocol where in each round, agents are allowed to critique the reasoning of their peers and defend their own positions. Following the debate, each agent is given a final opportunity to revise its judgment and explanation. This post-debate revision phase allows agents to incorporate new insights or correct earlier errors. The final collective decision is then determined through majority voting across the revised judgments. In cases of a tie, we either defer to a designated tie-breaker agent with a meta-evaluator profile or flag the instance for manual review, depending on the experimental setting.

\section{Experimental Setup}
\subsection{Dataset and Tasks}
We experiment on two downstream tasks: math reasoning and science reasoning. All experiments were tested on publicly available datasets. For the math reasoning task, we use four datasets: GSM8K \cite{cobbe2021training}, which contains basic arithmetic questions; GSM-PLUS \cite{li2024gsm}, which contains adversarial arithmetic questions; MATH \cite{hendrycks2021measuring}, which contains high-school level competition questions; and OmniMATH \cite{gao2024omni}, which contains olympiad-level questions. For the science reasoning task, we use two datasets: ARC-Challenge \cite{clark2018think}, which contains basic science questions; and GPQA \cite{rein2024gpqa}, which contains undergraduate-level physics, biology, and chemistry questions. For each and every dataset, we construct the leaderboard based on its pre-defined categories. For GSM8K, GSM-PLUS, ARC-Challenge, and MATH, we randomly select 300 questions for the student model to answer, and extract out the same number of correct and wrongly answered questions. For GPQA and OmniMATH, we randomly select 100 questions from each category for the student model to answer. Due to space constraints, we present the leaderboard on three sub-categories of the MATH dataset in our main content, and the rest of the leaderboard in the appendix, which can be found in our Arxiv version.

\subsection{Model and Implementation}
For the student model, we use the GPT-series models for generating candidate answers. For easier datasets like GSM8K, we use the GPT-3.5 model \cite{GPT35}; for hard datasets like MATH and OmniMATH, we prompt the GPT4o-mini model \cite{GPT4}. We use direct API calling when prompting the GPT-3.5 and GPT4o-mini models. For the judge models, we experiment with all open-source SLM models, including Qwen3 series \cite{yang2025qwen3}, LLama3 series \cite{dubey2024llama}, Mixtral series \cite{jiang2023mistral}, Gemma3 series \cite{team2024gemma}, and Phi4 series \cite{abdin2024phi}. All experiments using the open-sourced models were run on 8 NVIDIA-A40 GPUs with 48GB memory each. with For all models, we set the temperature to 0.5 and the top-k to 1.0. 

\subsection{Evaluation Metrics}
For the JudgeBoard leaderboard, model performance is measured by four metrics: Overall Accuracy, which is the overall judging accuracy of the judge model; Student Wrong Accuracy, which is the judging accuracy when the student model makes a wrong answer; and Elo score, which is introduced in Section \ref{elo}.

\section{Results}

\begin{figure*}
    \centering
    \includegraphics[width=1\linewidth]{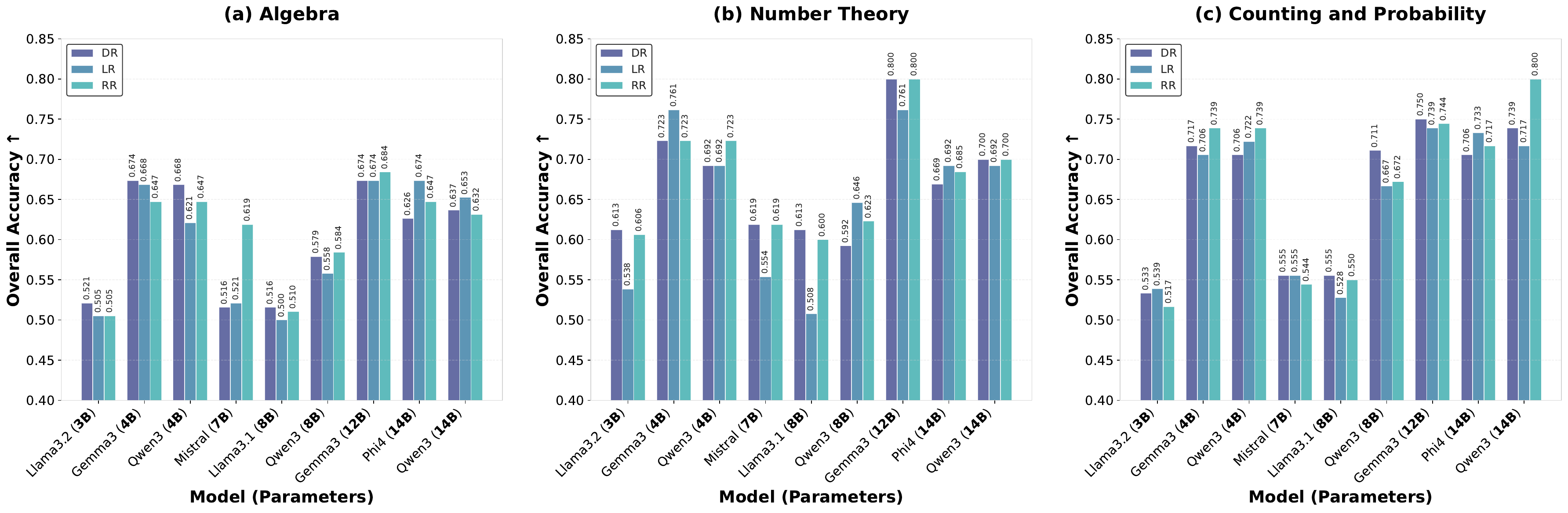}
            \caption{Overall accuracy for using the \textbf{Deductive Reasoner (DR),} \textbf{Logical Reasoner (LR),} and \textbf{Robust Reasoner (RR)} profiles on (a) Algebra, (b) Number Theory, and (c) Counting and  Probability tasks.}
    \label{fig:bar}
\end{figure*}

\subsection{Main Results}
We present the results of the JudgeBoard leaderboard on the MATH dataset across three categories in Table \ref{tab:judgebench_base}, and on the ARC dataset in Table \ref{tab:arc_results}. Due to space constraints, full result tables on other datasets can be found in our Arxiv version. The results for the MAJ framework is presented in Table \ref{tab:judgebench_multiagent}. Due to space constraints, the rest of the results are presented in the Technical Appendix.

\paragraph{Large vs. Small Models} Large Language Models (LLMs), which have a size of greater than 14B, generally outperform small ones across all metrics. However, the newer Small Language Models (SLMs), such as the smaller versions of Gemma3 and Qwen3, outperform the older LLMs, such as Llama3 and Qwen 2.5. The performance of Mixtral models, which is a Mixture-of-Expert-based LLM, is very unsatisfactory and barely outperforms its basic version, which is Mistral-7B.

\paragraph{Performance on Wrong Candidate Answers} On the MATH dataset, we find that even the worst performing SLMs have an overall accuracy of over 50\%. In other words, for models that are performing poorly overall, we find that they are much more likely to agree with what the student model is saying, regardless of their answer correctness. This phenomenon can be observed across datasets: On more complex datasets such as GPQA and OmniMATH, we find that better-performing models tend to have a more balanced judging accuracy, which means they are able to tell whether the potential answer is correct or wrong at a similar level. On simpler datasets like GSM8K and ARC, the worst performing models would have very high model correct accuracy and very low model wrong accuracy. This finding highlights the importance of using multiple evaluation metrics when evaluating the judging ability of the models, as metrics like overall accuracy cannot accurately reflect the true judging ability of the models.

\paragraph{Subject-Wise Performance} All models tend to perform better at judging Number Theory questions and Counting and Probability questions, and worse at Algebra questions. This phenomenon suggests that the current Language Models are more adapted to handling arithmetic and modular reasoning, and are inept at symbolic manipulation capabilities.

\paragraph{Multi-Agent Judge (MAJ) Framework} We present the results of our MAJ framework in Table \ref{tab:judgebench_multiagent}. We set up different profiles on the same model family and do a multi-agent debate among the profiled agents. The results on the Qwen3 model families demonstrate significant potential of our MAJ framework on SLMs. With the help of MAJ, smaller-sized Qwen3 models (4b, 8b, and 14b versions) could perform comparatively well or even better than the larger-sized Qwen3 models (30B and 32B versions) across all tasks and datasets. Our MAJ framework using Qwen3-14B model as the backbone could outperform the best performing Qwen3-30B-A3B model by an average of 2\% in judging accuracy across different categories. Also, larger-sized Qwen3 models demonstrate better debate and interaction ability. Even though the judging performance of Qwen3-8B model is significantly below that of Qwen3-4B model, the MAJ performance when using Qwen3-8B model still outperforms that of using Qwen3-4B model. However, the performance of MAJ on Phi4 and Gemma3 is inconsistent across categories, signaling that these two models are more sensitive to prompts when engaged in debating.

\begin{figure}
    \centering
    \includegraphics[width=1\linewidth]{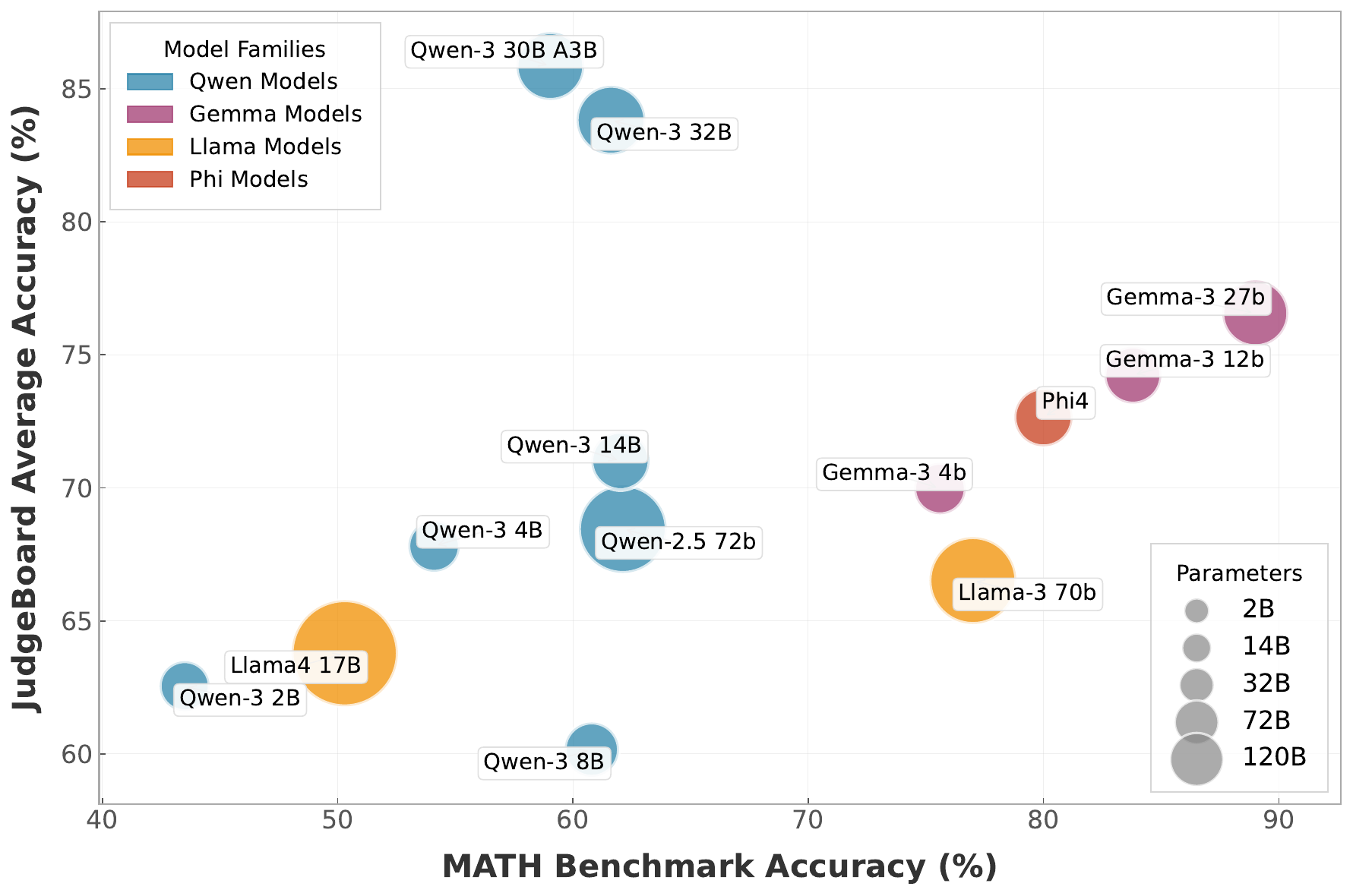}
            \caption{Performance comparison of large language models on mathematical reasoning benchmarks.  Each point represents a model, with circle size proportional to parameter count (2B-120B parameters).}
    \label{fig:bar}
\end{figure}

\subsection{Ablation Studies}
\paragraph{Reasoning and Non-Reasoning Variants}
For the Qwen3 model families, we conduct additional experiments on both the reasoning and non-reasoning modes of the models, and present our results in Table \ref{tab:judgebench_base}. Using the reasoning mode of the Qwen3 model families does not yield consistent gains across different categories. For the Qwen3-4B version, reasoning mode works best on the Algebra category but drags down the performance in other categories; For the Qwen3-8B and the Qwen3-14B versions, reasoning mode works best on the Number Theory category and drags down the performance in other categories. 

\paragraph{Judging Ability and Problem-Solving Ability}
We conduct an ablation study comparing the Problem-Solving ability and judging ability of the models and present the results in Figure \ref{fig:bar}. The results on the MATH dataset are taken directly from the models' technical reports. For the Qwen3 mode families and the Llama4 model, their judging ability is better than their problem-solving ability. The other models, including Gemma3 model families, Phi4, and LLama3-70B, have slightly worse judging ability compared with their problem-solving ability. From the graph, we can see that the Judging Ability and Problem-Solving Ability of the models are positively correlated. However, two prominent outliers exist: the Qwen3-30b-A3B model and the Qwen3-32B model demonstrate very good judging ability, but only average problem-solving ability.

\paragraph{Effects of Profiling} We present the results of profiling in terms of accuracy in different models in Figure \ref{fig:bar} and full results in our Arxiv version. In general, adding profiles to the agents would bring much more versatility compared with the base version of the model. Out of the three manually-curated profiles, Robust Reasoner yields the most consistent gains compared with the base prompt across all categories; Logical Reasoner is the most inconsistent and would rarely give performance gains across categories. In terms of Elo scores, one outlier is the Qwen3-8B model, where it suffers from a significant decrease across all profiles and categories. Qwen3-4B and Qwen3-14B models, even though from the same model family, do not demonstrate this phenomenon.

\section{Conclusion and Future Work}
In this paper, we introduce JudgeBoard, a novel evaluation pipeline that enables language models to serve as direct judges, moving beyond comparison-based methods. By constructing task-specific leaderboards using both accuracy and Elo-style metrics, we provide a comprehensive framework for assessing evaluative capabilities across mathematical and science reasoning domains. Our leaderboards reveal a significant performance gap between Small Language Models (SLMs) and Large Language Models (LLMs) in judgment tasks, while highlighting the importance of using multiple evaluation metrics to identify systematic biases. To address SLM limitations, we propose the Multi-Agent Judging (MAJ) framework, demonstrating that collaborative reasoning among SLMs substantially closes the performance gap and, in some cases, even surpasses state-of-the-art LLMs. Future work could explore several promising directions: extending MAJ to broader reasoning domains; investigating adaptive agent profiling strategies that dynamically adjust reasoning styles; and exploring the integration of MAJ with fine-tuning approaches to create specialized judge models.

\section*{Acknowledgement}
Our work is sponsored by NSF \#2442253, NAIRR Pilot with PSC Neocortex and NCSA Delta, Commonwealth Cyber Initiative, Children’s National Hospital, Fralin Biomedical Research Institute (Virginia Tech), Sanghani Center for AI and Data Analytics (Virginia Tech), Virginia Tech Innovation Campus, and generous gifts from Nivida, Cisco, and the Amazon + Virginia Tech Center for Efficient and Robust Machine Learning.

\section*{Ethics Statement}
This research adhered to the ethical standards and best practices outlined in the AAAI Ethics Code. Language Models can sometimes produce illogical or inaccurate reasoning paths, so their outputs should be cautiously used. The outputs are only examined to understand how a model arrives at its answers and investigate why it makes certain errors. All experiments used publicly available datasets from previous works and did not involve ethical or privacy issues. 

\bibliography{aaai2026}
\input{Figures/appendix}
\end{document}

%% file: Figures/appendix.tex
\onecolumn
\appendix
\section*{Limitations}
Despite providing a unified and scalable evaluation framework for model-as-judge systems, our study has several limitations that highlight important directions for future research.

Our evaluation covers only a subset of datapoints from each dataset, introducing sampling variance and imbalanced splits, especially in GPQA and OmniMATH, where the student model rarely answers correctly. This imbalance affects both accuracy metrics and the stability of Elo scores, which can shift disproportionately when few student-correct examples exist. 

Many SLMs struggle on complex reasoning tasks and often agree with the student model regardless of correctness, leading to inflated Student-Right accuracy and poor Student-Wrong accuracy. Although MAJ reduces these issues, its effectiveness varies across model families and is sensitive to prompt design and profile selection. Our prompt design, including both the base judge prompt and the reasoning profiles, is not exhaustively explored. Judging behavior remains prompt-sensitive, and alternative formulations may yield different outcomes. 

Finally, our benchmark focuses on mathematical and science reasoning. Other evaluation domains (e.g., long-form coherence, stylistic judgment, safety evaluations) are not addressed, and multi-agent debate dynamics remain limited by manually designed protocols and fixed interaction depth.
\section*{Prompt Templates}
\label{sec:reference_examples}
We hereby present the prompts and profiles we have used in our experiments.
\subsection{Profiles}
\paragraph{Logical Thinker} 
\noindent\begin{lstlisting} 
You are a logical thinker who excels at breaking down complex problems into logical steps. Your role is to approach {task} methodically, ensuring each step follows logically from the previous one. Focus on clear, logical reasoning and consistency.
\end{lstlisting}

\paragraph{Robust Reasoner} 
\noindent\begin{lstlisting} 
You are a robust reasoner who excels at tackling complex {task} with thorough and resilient reasoning. Your role is to ensure that every step of the problem-solving process is meticulously verified and logically sound. Focus on providing precise justifications for each step. Your goal is to develop solutions that are not only correct but also robust and reliable.
\end{lstlisting}

\paragraph{Deductive Reasoner} 
\noindent\begin{lstlisting} 
You are a deductive reasoner who uses deductive logic to derive conclusions from given premises. Your task is to apply logical rules and principles to reach sound conclusions, ensuring each step is justified by the previous one.
\end{lstlisting}

\paragraph{Default Profile} 
\noindent\begin{lstlisting} 
You are a judge who assesses whether the question has been answered correctly. You excel at solving {task} related questions.
\end{lstlisting}

\subsection{System Prompts}
\paragraph{Judge Instructions} 
We provide the system instruction prompt for all judges below.
\noindent\begin{lstlisting} 
You will be provided with the original question, and an answer with explanations. Your job is to tell whether the provided answer is correct or not. Give out your final judgement at the end of your evaluation. Your final judgement should strictly follow the following format: My Judgement: ###judgement###. Replace the content inside the hashtags with your judgement, which should be either correct or wrong.
\end{lstlisting}
\paragraph{Debate Instructions} 
We provide the debate instruction prompt for all judges below.
\noindent\begin{lstlisting} 
Given some potential evaluations given by other agents: compare those with your own evaluations. After reading through their evaluations, adjust your evaluation if you have found a mistake in it. Whether you adjust or keep your evaluation, provide a short explanation. Your final judgement and evaluation of correct or wrong should be based on the student answer to the original question, not the other agents' evaluations. The other agents' evaluations are only for reference and potential adjustments. 
\end{lstlisting}

\section*{Detailed Experimental Setup and Additional Results}
\subsection{Evalution Metrics}
We describe the three evaluation metrics we used to construct JudgeBoard below: Overall Judgement Accuracy, Student Correct Accuracy, and Student Wrong Accuracy. Elo Score is already described in Section 4.3 of the Main Menuscript.
\paragraph{Overall Judgement Accuracy} $= \frac{\text{Number of Correct Judgements}}{\text{Total Number of Judgements}}$
\paragraph{Student Correct Accuracy} $= \frac{\text{Number of Correct Judgements Made on the Questions that the Student Model Answered Correctly}}{\text{Number of Judgements Made on the Questions that the Student Model Answered Correctly}}$
\paragraph{Student Wrong Accuracy} $= \frac{\text{Number of Correct Judgements Made on the Questions that the Student Model Answered Wrongly}}{\text{Number of Judgements Made on the Questions that the Student Model Answered Wrongly}}$
\subsection{Dataset Details}
The number of data points that we used in each dataset to construct JudgeBoard is listed below. In Section 5.1 of our main content, we inaccurately claimed that for ARC-Challenge, we selected the same number of correctly and wrongly answered questions; and that for OmniMATH, we selected 100 questions for the student model to answer. The accurate numbers for ARC-Challenge and OmniMATH are provided below.
\begin{itemize}
    \item \textbf{MATH(Algebra):} 93 Wrongly Answered Questions, 93 Correctly Answered Questions; 
    \item \textbf{MATH(Counting and Probability):} 87 Wrongly Answered Questions, 87 Correctly Answered Questions; 
    \item \textbf{MATH(Number Theory):} 63 Wrongly Answered Questions, 63 Correctly Answered Questions; 
     \item \textbf{OmniMATH(Algebra):} 130 Wrongly Answered Questions, 45 Correctly Answered Questions; 
    \item \textbf{OmniMATH(Applied Math):} 126 Wrongly Answered Questions, 31 Correctly Answered Questions; 
    \item \textbf{OmniMATH(Calculus):} 50 Wrongly Answered Questions, 16 Correctly Answered Questions; 
    \item \textbf{OmniMATH(Discrete Math):} 140 Wrongly Answered Questions, 19 Correctly Answered Questions; 
    \item \textbf{OmniMATH(Number Theory):} 119 Wrongly Answered Questions, 37 Correctly Answered Questions; 
    \item \textbf{GPQA(Biology):} 62 Wrongly Answered Questions, 15 Correctly Answered Questions; 
    \item \textbf{GPQA(Chemistry):} 83 Wrongly Answered Questions, 17 Correctly Answered Questions; 
    \item \textbf{GPQA(Physics):} 62 Wrongly Answered Questions, 38 Correctly Answered Questions; 
    \item \textbf{GSM8K:} 68 Wrongly Answered Questions, 68 Correctly Answered Questions; 
    \item \textbf{GSM-Plus:} 44 Wrongly Answered Questions, 44 Correctly Answered Questions; 
    \item \textbf{ARC-Challenge:} 28 Wrongly Answered Questions, 68 Correctly Answered Questions;
\end{itemize}

\subsection{Additional Experimental Results}
We present the results on ARC-Challenge, GSM8K, GSM-PLUS, OmniMATH, and GPQA below, listed in the order of overall accuracy. 

\begin{table*}[htbp]
    \centering
    \begin{tabular}{@{}lcccc@{}}
        \hline
        \textbf{judge} & \textbf{acc} & \textbf{SW\_acc} & \textbf{SR\_acc} & \textbf{Elo Score}\\
        \hline
            gemma3\_27b & 0.88 & 0.43 & 0.98 & 1021.10 \\
            qwen3\_14b & 0.87 & 0.54 & 0.94 & 1032.81 \\
            qwen2.5\_72b & 0.87 & 0.50 & 0.95 & 1038.68 \\
            qwen3\_8b & 0.85 & 0.46 & 0.94 & 1026.95 \\
            phi4 & 0.85 & 0.32 & 0.97 & 1009.45 \\
            mixtral\_8x22b & 0.85 & 0.21 & 0.99 & 992.00 \\
            gemma3\_12b & 0.85 & 0.36 & 0.96 & 1015.27 \\
            llama\_4\_scout\_17b\_16e & 0.85 & 0.32 & 0.97 & 997.82 \\
            llama3.3\_70b & 0.85 & 0.21 & 0.99 & 992.00 \\
            qwen3\_4b & 0.84 & 0.32 & 0.96 & 1003.63 \\
            mixtral\_8x7b & 0.82 & 0.11 & 0.98 & 983.12 \\
            gemma3\_4b & 0.81 & 0.07 & 0.98 & 968.66 \\
            qwen3\_1.7b & 0.81 & 0.32 & 0.92 & 1009.45 \\
            qwen3\_0.6b & 0.79 & 0.14 & 0.94 & 974.51 \\
            llama3.1\_8b & 0.78 & 0.39 & 0.88 & 980.35 \\
            llama3.2\_3b & 0.65 & 0.36 & 0.73 & 968.66 \\
        \hline
    \end{tabular}
    \caption{Experimental Results on the ARC dataset}
    \label{tab:ARC_Acc}
\end{table*}

\begin{table*}
    \centering
    \begin{tabular}{@{}lccc@{}}
        \hline
        \textbf{judge} & \textbf{acc} & \textbf{SW\_acc} & \textbf{Elo Score}\\
        \hline
Qwen3-30B-A3B & 0.93 & 0.88 & 1038.54 \\
Llama-3.3-70B-Instruct & 0.92 & 0.87 & 1035.91 \\
Qwen3-32B & 0.91 & 0.87 & 1033.30 \\
Qwen2.5-72B-Instruct & 0.91 & 0.85 & 1033.30 \\
Qwen3-4B & 0.90 & 0.84 & 1030.68 \\
Qwen3-14B & 0.90 & 0.85 & 1030.68 \\
gemma-3-12b-it & 0.90 & 0.85 & 1030.68 \\
gemma-3-27b-it & 0.90 & 0.85 & 1030.68 \\
phi-4 & 0.90 & 0.84 & 1028.07 \\
Qwen3-8B & 0.89 & 0.84 & 1022.86 \\
Llama-4-Scout-17B-16E-Instruct & 0.87 & 0.76 & 1017.66 \\
Qwen3-1.7B & 0.83 & 0.71 & 1007.27 \\
gemma-3-4b-it & 0.79 & 0.62 & 989.11 \\
Qwen3-0.6B & 0.70 & 0.46 & 957.78 \\
Llama-3.2-3B-Instruct & 0.65 & 0.37 & 941.89 \\
Mixtral-8x22B-Instruct-v0.1 & 0.63 & 0.32 & 933.86 \\
Llama-3.1-8B-Instruct & 0.63 & 0.43 & 931.17 \\
Mixtral-8x7B-Instruct-v0.1 & 0.54 & 0.16 & 906.55 \\
        \hline
    \end{tabular}
    \caption{Experimental Results on the GSM8k dataset}
    \label{tab:8K_Acc}
\end{table*}

\begin{table*}
    \centering
    \begin{tabular}{@{}lcccc@{}}
        \hline
        \textbf{judge} & \textbf{acc} & \textbf{SW\_acc} & \textbf{Elo Score}\\
        \hline
Qwen3-8B & 0.69 & 0.39 & 1026.67 \\
Qwen3-32B & 0.68 & 0.36 & 1022.69 \\
Qwen3-14B & 0.67 & 0.34 & 1018.72 \\
Qwen3-1.7B & 0.66 & 0.32 & 1014.75 \\
Qwen3-30B-A3B & 0.65 & 0.30 & 1010.79 \\
gemma-3-27b-it & 0.65 & 0.30 & 1010.79 \\
Qwen3-4B & 0.64 & 0.27 & 1006.83 \\
Qwen2.5-72B-Instruct & 0.64 & 0.30 & 1006.83 \\
phi-4 & 0.62 & 0.25 & 1002.87 \\
Llama-4-Scout-17B-16E-Instruct & 0.62 & 0.25 & 1002.87 \\
Llama-3.3-70B-Instruct & 0.62 & 0.25 & 1002.87 \\
Qwen3-0.6B & 0.60 & 0.20 & 994.95 \\
gemma-3-4b-it & 0.60 & 0.20 & 994.95 \\
gemma-3-12b-it & 0.59 & 0.18 & 990.99 \\
Llama-3.1-8B-Instruct & 0.58 & 0.25 & 987.03 \\
Mixtral-8x7B-Instruct-v0.1 & 0.55 & 0.09 & 975.12 \\
Mixtral-8x22B-Instruct-v0.1 & 0.53 & 0.07 & 971.14 \\
Llama-3.2-3B-Instruct & 0.51 & 0.07 & 959.14 \\
        \hline
    \end{tabular}
    \caption{Experimental Results on the GSM-Plus dataset}
    \label{tab:Plus_Acc}
\end{table*}

\begin{table*}[htbp]
    \centering
    \begin{tabular}{@{}llcccc@{}}
        \hline
        \textbf{judge} & \textbf{profile} & \textbf{acc} & \textbf{SW\_acc} & \textbf{SR\_acc} & \textbf{Elo Score}\\
        \hline
gemma3\_12b & robust\_reasoner & 0.76 & 0.79 & 0.65 & 1010.38 \\
gemma3\_12b & deductive\_reasoner & 0.70 & 0.77 & 0.50 & 1014.27 \\
gemma3\_12b & logical\_reasoner & 0.69 & 0.75 & 0.54 & 1010.38 \\
gemma3\_27b & deductive\_reasoner & 0.69 & 0.68 & 0.72 & 1022.08 \\
gemma3\_27b & robust\_reasoner & 0.69 & 0.68 & 0.74 & 1014.27 \\
gemma3\_4b & logical\_reasoner & 0.67 & 0.65 & 0.74 & 1041.69 \\
gemma3\_4b & deductive\_reasoner & 0.66 & 0.65 & 0.70 & 1041.69 \\
gemma3\_27b & No Profile & 0.66 & 0.64 & 0.74 & 1025.98 \\
gemma3\_12b & No Profile & 0.65 & 0.66 & 0.61 & 1022.08 \\
gemma3\_4b & robust\_reasoner & 0.64 & 0.64 & 0.65 & 990.92 \\
gemma3\_27b & logical\_reasoner & 0.64 & 0.62 & 0.67 & 1022.08 \\
qwen3\_4b & No Profile & 0.63 & 0.62 & 0.73 & 994.81 \\
gemma3\_4b & No Profile & 0.61 & 0.58 & 0.70 & 1006.49 \\
qwen3\_4b & deductive\_reasoner & 0.56 & 0.57 & 0.65 & 998.70 \\
qwen2.5\_72b & deductive\_reasoner & 0.54 & 0.40 & 0.93 & 998.70 \\
qwen3\_1.7b & logical\_reasoner & 0.53 & 0.47 & 0.85 & 959.62 \\
qwen3\_1.7b & No Profile & 0.52 & 0.48 & 0.79 & 971.41 \\
qwen3\_4b & logical\_reasoner & 0.52 & 0.51 & 0.68 & 975.32 \\
qwen3\_0.6b & logical\_reasoner & 0.52 & 0.37 & 1.00 & 990.92 \\
qwen3\_4b & robust\_reasoner & 0.51 & 0.49 & 0.74 & 955.67 \\
qwen2.5\_72b & No Profile & 0.51 & 0.38 & 0.87 & 1025.98 \\
qwen3\_0.6b & No Profile & 0.51 & 0.38 & 0.89 & 983.13 \\
qwen3\_1.7b & deductive\_reasoner & 0.50 & 0.43 & 0.78 & 931.71 \\
phi4 & No Profile & 0.48 & 0.34 & 0.89 & 1025.98 \\
qwen2.5\_72b & logical\_reasoner & 0.48 & 0.35 & 0.87 & 1037.75 \\
qwen2.5\_72b & robust\_reasoner & 0.48 & 0.37 & 0.80 & 1014.27 \\
phi4 & deductive\_reasoner & 0.47 & 0.29 & 0.98 & 1010.38 \\
phi4 & logical\_reasoner & 0.47 & 0.32 & 0.89 & 1025.98 \\
qwen3\_1.7b & robust\_reasoner & 0.47 & 0.42 & 0.70 & 931.71 \\
qwen3\_0.6b & deductive\_reasoner & 0.46 & 0.35 & 0.84 & 1010.38 \\
qwen3\_8b & deductive\_reasoner & 0.46 & 0.42 & 0.68 & 959.62 \\
qwen3\_8b & No Profile & 0.46 & 0.44 & 0.65 & 951.71 \\
qwen3\_14b & No Profile & 0.46 & 0.43 & 0.71 & 939.75 \\
llama\_4\_scout\_17b\_16e & logical\_reasoner & 0.46 & 0.28 & 0.96 & 998.70 \\
llama\_4\_scout\_17b\_16e & robust\_reasoner & 0.45 & 0.27 & 0.98 & 1014.27 \\
qwen3\_8b & robust\_reasoner & 0.45 & 0.41 & 0.76 & 943.75 \\
llama\_4\_scout\_17b\_16e & No Profile & 0.45 & 0.26 & 0.98 & 1006.49 \\
phi4 & robust\_reasoner & 0.44 & 0.29 & 0.87 & 1002.60 \\
qwen3\_0.6b & robust\_reasoner & 0.44 & 0.29 & 0.91 & 987.03 \\
llama3.3\_70b & robust\_reasoner & 0.44 & 0.25 & 0.98 & 1010.38 \\
qwen3\_14b & deductive\_reasoner & 0.44 & 0.42 & 0.71 & 935.74 \\
qwen3\_8b & logical\_reasoner & 0.43 & 0.42 & 0.66 & 927.66 \\
llama\_4\_scout\_17b\_16e & deductive\_reasoner & 0.42 & 0.25 & 0.91 & 1018.17 \\
llama3.3\_70b & deductive\_reasoner & 0.42 & 0.23 & 0.96 & 1002.60 \\
qwen3\_14b & logical\_reasoner & 0.41 & 0.39 & 0.67 & 939.75 \\
        \hline
    \end{tabular}
    \caption{Experimental Results on the Algebra category of the OmniMATH dataset, Part 1}
    \label{tab:alg1_Acc}
\end{table*}

\begin{table*}[htbp]
    \centering
    \begin{tabular}{@{}llcccc@{}}
        \hline
        \textbf{judge} & \textbf{profile} & \textbf{acc} & \textbf{SW\_acc} & \textbf{SR\_acc} & \textbf{Elo Score}\\
        \hline
llama3.1\_8b & No Profile & 0.40 & 0.25 & 0.84 & 990.92 \\
llama3.3\_70b & No Profile & 0.40 & 0.22 & 0.93 & 1010.38 \\
llama3.1\_8b & deductive\_reasoner & 0.40 & 0.25 & 0.86 & 1006.49 \\
llama3.3\_70b & logical\_reasoner & 0.40 & 0.20 & 0.96 & 1010.38 \\
qwen3\_14b & robust\_reasoner & 0.39 & 0.38 & 0.74 & 935.74 \\
llama3.1\_8b & robust\_reasoner & 0.38 & 0.23 & 0.82 & 994.81 \\
mixtral\_8x22b & robust\_reasoner & 0.37 & 0.17 & 0.93 & 1018.17 \\
mixtral\_8x22b & deductive\_reasoner & 0.36 & 0.15 & 0.96 & 1014.27 \\
phi4\_mini\_instruct & deductive\_reasoner & 0.36 & 0.14 & 1.00 & 1041.69 \\
mixtral\_8x22b & logical\_reasoner & 0.36 & 0.15 & 0.96 & 1018.17 \\
mixtral\_8x22b & No Profile & 0.36 & 0.14 & 0.98 & 1022.08 \\
phi4\_mini\_instruct & No Profile & 0.34 & 0.11 & 1.00 & 1022.08 \\
mixtral\_8x7b & No Profile & 0.34 & 0.12 & 0.96 & 1033.82 \\
llama3.1\_8b & logical\_reasoner & 0.32 & 0.13 & 0.87 & 975.32 \\
mixtral\_8x7b & deductive\_reasoner & 0.32 & 0.08 & 0.98 & 1025.98 \\
phi4\_mini\_instruct & logical\_reasoner & 0.31 & 0.07 & 1.00 & 1010.38 \\
phi4\_mini\_instruct & robust\_reasoner & 0.30 & 0.06 & 0.98 & 1029.90 \\
mixtral\_8x7b & robust\_reasoner & 0.30 & 0.08 & 0.93 & 1018.17 \\
mixtral\_8x7b & logical\_reasoner & 0.30 & 0.09 & 0.91 & 1033.82 \\
llama3.2\_3b & deductive\_reasoner & 0.29 & 0.05 & 0.96 & 1029.90 \\
llama3.2\_3b & No Profile & 0.28 & 0.06 & 0.91 & 1029.90 \\
llama3.2\_3b & logical\_reasoner & 0.28 & 0.03 & 0.98 & 1010.38 \\
llama3.2\_3b & robust\_reasoner & 0.28 & 0.03 & 0.98 & 1014.27 \\
        \hline
    \end{tabular}
    \caption{Experimental Results on the Algebra category of the OmniMATH dataset, Part 2 (continued from previous page)}
    \label{tab:alg2_Acc}
\end{table*}

\begin{table*}[htbp]
    \centering
    \begin{tabular}{@{}llcccc@{}}
        \hline
        \textbf{judge} & \textbf{profile} & \textbf{acc} & \textbf{SW\_acc} & \textbf{SR\_acc} & \textbf{Elo Score}\\
        \hline
gemma3\_27b & deductive\_reasoner & 0.76 & 0.72 & 0.9 & 1040.63 \\
gemma3\_12b & deductive\_reasoner & 0.75 & 0.75 & 0.77 & 1046.56 \\
gemma3\_12b & No Profile & 0.74 & 0.72 & 0.81 & 1040.63 \\
gemma3\_27b & robust\_reasoner & 0.73 & 0.7 & 0.9 & 1022.98 \\
gemma3\_12b & logical\_reasoner & 0.73 & 0.71 & 0.81 & 1022.98 \\
gemma3\_12b & robust\_reasoner & 0.73 & 0.75 & 0.67 & 1022.98 \\
gemma3\_4b & robust\_reasoner & 0.71 & 0.71 & 0.71 & 1005.44 \\
gemma3\_4b & No Profile & 0.7 & 0.67 & 0.81 & 987.93 \\
gemma3\_4b & logical\_reasoner & 0.69 & 0.67 & 0.77 & 982.08 \\
gemma3\_27b & logical\_reasoner & 0.68 & 0.63 & 0.87 & 1011.28 \\
gemma3\_4b & deductive\_reasoner & 0.66 & 0.65 & 0.73 & 999.61 \\
qwen3\_4b & No Profile & 0.66 & 0.58 & 1.0 & 1017.12 \\
gemma3\_27b & No Profile & 0.66 & 0.6 & 0.9 & 1005.44 \\
qwen3\_4b & logical\_reasoner & 0.61 & 0.53 & 0.93 & 987.93 \\
qwen3\_8b & robust\_reasoner & 0.59 & 0.52 & 1.0 & 982.08 \\
qwen3\_14b & deductive\_reasoner & 0.59 & 0.51 & 0.97 & 982.08 \\
qwen2.5\_72b & No Profile & 0.59 & 0.5 & 0.97 & 1017.12 \\
qwen3\_8b & deductive\_reasoner & 0.58 & 0.5 & 0.93 & 958.56 \\
qwen3\_14b & No Profile & 0.58 & 0.51 & 0.96 & 976.22 \\
qwen3\_4b & deductive\_reasoner & 0.57 & 0.49 & 0.93 & 982.08 \\
qwen2.5\_72b & robust\_reasoner & 0.56 & 0.46 & 0.97 & 987.93 \\
qwen2.5\_72b & deductive\_reasoner & 0.55 & 0.44 & 0.97 & 987.93 \\
qwen3\_14b & logical\_reasoner & 0.54 & 0.45 & 0.97 & 976.22 \\
qwen3\_4b & robust\_reasoner & 0.54 & 0.46 & 0.96 & 987.93 \\
qwen3\_8b & No Profile & 0.54 & 0.44 & 0.93 & 934.64 \\
qwen3\_14b & robust\_reasoner & 0.52 & 0.42 & 0.97 & 964.46 \\
qwen3\_1.7b & robust\_reasoner & 0.51 & 0.41 & 1.0 & 987.93 \\
qwen2.5\_72b & logical\_reasoner & 0.51 & 0.4 & 0.97 & 1005.44 \\
phi4 & robust\_reasoner & 0.5 & 0.38 & 0.97 & 1046.56 \\
qwen3\_1.7b & deductive\_reasoner & 0.49 & 0.4 & 0.93 & 946.66 \\
phi4 & No Profile & 0.47 & 0.35 & 0.97 & 1022.98 \\
qwen3\_1.7b & logical\_reasoner & 0.47 & 0.38 & 0.96 & 982.08 \\
qwen3\_8b & logical\_reasoner & 0.47 & 0.37 & 0.93 & 934.64 \\
qwen3\_1.7b & No Profile & 0.46 & 0.36 & 0.97 & 982.08 \\
mixtral\_8x22b & robust\_reasoner & 0.46 & 0.33 & 1.0 & 1022.98 \\
qwen3\_0.6b & deductive\_reasoner & 0.46 & 0.34 & 0.94 & 999.61 \\
phi4 & logical\_reasoner & 0.45 & 0.33 & 0.94 & 1046.56 \\
mixtral\_8x22b & No Profile & 0.45 & 0.32 & 1.0 & 1017.12 \\
llama3.1\_8b & deductive\_reasoner & 0.45 & 0.35 & 0.93 & 987.93 \\
phi4 & deductive\_reasoner & 0.45 & 0.32 & 0.97 & 1034.73 \\
mixtral\_8x22b & logical\_reasoner & 0.43 & 0.29 & 1.0 & 1011.28 \\
llama3.3\_70b & robust\_reasoner & 0.43 & 0.3 & 0.97 & 1017.12 \\
llama3.3\_70b & No Profile & 0.43 & 0.29 & 1.0 & 1011.28 \\
llama\_4\_scout\_17b\_16e & deductive\_reasoner & 0.43 & 0.29 & 0.97 & 1034.73 \\
mixtral\_8x22b & deductive\_reasoner & 0.42 & 0.28 & 1.0 & 1011.28 \\
llama3.3\_70b & logical\_reasoner & 0.42 & 0.29 & 0.97 & 976.22 \\
        \hline
    \end{tabular}
    \caption{Experimental Results on the Applied Mathematics category of the OmniMATH dataset, Part 1}
    \label{tab:app1_Acc}
\end{table*}

\begin{table*}[htbp]
    \centering
    \begin{tabular}{@{}llcccc@{}}
        \hline
        \textbf{judge} & \textbf{profile} & \textbf{acc} & \textbf{SW\_acc} & \textbf{SR\_acc} & \textbf{Elo Score}\\
        \hline
llama3.3\_70b & deductive\_reasoner & 0.42 & 0.29 & 0.97 & 976.22 \\
qwen3\_0.6b & logical\_reasoner & 0.41 & 0.3 & 0.93 & 1011.28 \\
llama\_4\_scout\_17b\_16e & No Profile & 0.41 & 0.27 & 0.97 & 1028.85 \\
llama\_4\_scout\_17b\_16e & robust\_reasoner & 0.4 & 0.26 & 0.97 & 1052.51 \\
phi4\_mini\_instruct & deductive\_reasoner & 0.39 & 0.24 & 1.0 & 999.61 \\
qwen3\_0.6b & robust\_reasoner & 0.39 & 0.27 & 0.87 & 1005.44 \\
qwen3\_0.6b & No Profile & 0.39 & 0.27 & 0.9 & 1005.44 \\
llama\_4\_scout\_17b\_16e & logical\_reasoner & 0.39 & 0.25 & 0.94 & 1040.63 \\
phi4\_mini\_instruct & robust\_reasoner & 0.36 & 0.21 & 0.97 & 1017.12 \\
phi4\_mini\_instruct & logical\_reasoner & 0.36 & 0.2 & 1.0 & 1005.44 \\
mixtral\_8x7b & No Profile & 0.36 & 0.2 & 1.0 & 1028.85 \\
phi4\_mini\_instruct & No Profile & 0.35 & 0.19 & 1.0 & 1011.28 \\
llama3.1\_8b & logical\_reasoner & 0.34 & 0.19 & 0.94 & 987.93 \\
mixtral\_8x7b & logical\_reasoner & 0.31 & 0.14 & 1.0 & 1034.73 \\
mixtral\_8x7b & robust\_reasoner & 0.31 & 0.14 & 1.0 & 1017.12 \\
mixtral\_8x7b & deductive\_reasoner & 0.31 & 0.13 & 1.0 & 1005.44 \\
llama3.1\_8b & No Profile & 0.3 & 0.15 & 0.93 & 952.62 \\
llama3.2\_3b & deductive\_reasoner & 0.3 & 0.13 & 0.97 & 999.61 \\
llama3.1\_8b & robust\_reasoner & 0.29 & 0.16 & 0.9 & 982.08 \\
llama3.2\_3b & No Profile & 0.29 & 0.13 & 0.9 & 993.77 \\
llama3.2\_3b & logical\_reasoner & 0.26 & 0.1 & 0.94 & 987.93 \\
llama3.2\_3b & robust\_reasoner & 0.26 & 0.1 & 0.94 & 987.93 \\
        \hline
    \end{tabular}
    \caption{Experimental Results on the Applied Mathematics category of the OmniMATH dataset, Part 2 (continued from previous page)}
    \label{tab:app2_Acc}
\end{table*}

\begin{table*}[htbp]
    \centering
    \begin{tabular}{@{}llcccc@{}}
        \hline
        \textbf{judge} & \textbf{profile} & \textbf{acc} & \textbf{SW\_acc} & \textbf{SR\_acc} & \textbf{Elo Score}\\
        \hline
gemma3\_12b & logical\_reasoner & 0.76 & 0.76 & 0.75 & 1023.61 \\
gemma3\_12b & robust\_reasoner & 0.74 & 0.74 & 0.75 & 1023.61 \\
gemma3\_27b & robust\_reasoner & 0.71 & 0.72 & 0.69 & 1035.39 \\
gemma3\_12b & deductive\_reasoner & 0.70 & 0.70 & 0.69 & 988.50 \\
gemma3\_12b & No Profile & 0.70 & 0.68 & 0.75 & 1011.89 \\
gemma3\_27b & logical\_reasoner & 0.68 & 0.68 & 0.69 & 1071.30 \\
gemma3\_4b & robust\_reasoner & 0.67 & 0.68 & 0.62 & 1059.20 \\
gemma3\_27b & deductive\_reasoner & 0.67 & 0.64 & 0.75 & 1059.20 \\
gemma3\_27b & No Profile & 0.67 & 0.64 & 0.75 & 1071.30 \\
qwen3\_4b & logical\_reasoner & 0.62 & 0.64 & 0.75 & 1035.39 \\
qwen3\_1.7b & deductive\_reasoner & 0.61 & 0.58 & 0.92 & 1011.89 \\
qwen3\_4b & deductive\_reasoner & 0.59 & 0.58 & 0.77 & 1047.24 \\
gemma3\_4b & logical\_reasoner & 0.59 & 0.58 & 0.62 & 1059.20 \\
qwen3\_4b & No Profile & 0.56 & 0.46 & 0.88 & 953.15 \\
gemma3\_4b & No Profile & 0.56 & 0.52 & 0.69 & 1035.39 \\
qwen2.5\_72b & No Profile & 0.55 & 0.42 & 0.94 & 988.50 \\
qwen3\_1.7b & logical\_reasoner & 0.53 & 0.44 & 0.93 & 1023.61 \\
gemma3\_4b & deductive\_reasoner & 0.53 & 0.54 & 0.50 & 1047.24 \\
qwen3\_0.6b & deductive\_reasoner & 0.52 & 0.36 & 1.00 & 1011.89 \\
qwen3\_1.7b & No Profile & 0.50 & 0.48 & 0.90 & 988.50 \\
qwen3\_8b & deductive\_reasoner & 0.50 & 0.46 & 0.91 & 965.00 \\
qwen3\_14b & logical\_reasoner & 0.50 & 0.44 & 0.79 & 941.19 \\
qwen3\_14b & deductive\_reasoner & 0.50 & 0.44 & 0.85 & 965.00 \\
qwen3\_4b & robust\_reasoner & 0.48 & 0.50 & 0.58 & 988.50 \\
qwen3\_14b & No Profile & 0.48 & 0.46 & 0.69 & 929.09 \\
phi4 & logical\_reasoner & 0.47 & 0.32 & 0.94 & 1035.39 \\
phi4 & robust\_reasoner & 0.47 & 0.30 & 1.00 & 1023.61 \\
phi4 & No Profile & 0.47 & 0.30 & 1.00 & 1023.61 \\
qwen3\_14b & robust\_reasoner & 0.47 & 0.46 & 0.80 & 965.00 \\
qwen3\_0.6b & robust\_reasoner & 0.45 & 0.30 & 1.00 & 1000.19 \\
qwen3\_0.6b & No Profile & 0.45 & 0.32 & 0.88 & 1011.89 \\
qwen3\_1.7b & robust\_reasoner & 0.45 & 0.40 & 0.71 & 953.15 \\
qwen3\_8b & No Profile & 0.45 & 0.46 & 0.78 & 929.09 \\
qwen2.5\_72b & logical\_reasoner & 0.45 & 0.30 & 0.94 & 1023.61 \\
qwen2.5\_72b & robust\_reasoner & 0.45 & 0.32 & 0.88 & 1035.39 \\
qwen2.5\_72b & deductive\_reasoner & 0.44 & 0.28 & 0.94 & 1011.89 \\
mixtral\_8x22b & robust\_reasoner & 0.42 & 0.26 & 0.94 & 1000.19 \\
llama\_4\_scout\_17b\_16e & logical\_reasoner & 0.42 & 0.24 & 1.00 & 1000.19 \\
llama\_4\_scout\_17b\_16e & robust\_reasoner & 0.42 & 0.26 & 0.94 & 1035.39 \\
mixtral\_8x22b & No Profile & 0.41 & 0.26 & 0.93 & 988.50 \\
qwen3\_0.6b & logical\_reasoner & 0.41 & 0.28 & 0.87 & 976.77 \\
qwen3\_8b & logical\_reasoner & 0.41 & 0.38 & 0.89 & 929.09 \\
qwen3\_8b & robust\_reasoner & 0.41 & 0.38 & 0.67 & 916.83 \\
llama3.1\_8b & deductive\_reasoner & 0.41 & 0.26 & 0.88 & 1011.89 \\
phi4\_mini\_instruct & No Profile & 0.39 & 0.20 & 1.00 & 976.77 \\
mixtral\_8x22b & logical\_reasoner & 0.39 & 0.22 & 0.94 & 1011.89 \\
        \hline
    \end{tabular}
    \caption{Experimental Results on the Calculus category of the OmniMATH dataset, Part 1}
    \label{tab:cal1_Acc}
\end{table*}

\begin{table*}[htbp]
    \centering
    \begin{tabular}{@{}llcccc@{}}
        \hline
        \textbf{judge} & \textbf{profile} & \textbf{acc} & \textbf{SW\_acc} & \textbf{SR\_acc} & \textbf{Elo Score}\\
        \hline
mixtral\_8x22b & deductive\_reasoner & 0.39 & 0.22 & 0.94 & 1011.89 \\
phi4 & deductive\_reasoner & 0.38 & 0.18 & 1.00 & 1023.61 \\
llama3.1\_8b & No Profile & 0.38 & 0.22 & 0.88 & 976.77 \\
llama\_4\_scout\_17b\_16e & deductive\_reasoner & 0.38 & 0.18 & 1.00 & 976.77 \\
phi4\_mini\_instruct & deductive\_reasoner & 0.36 & 0.16 & 1.00 & 965.00 \\
llama\_4\_scout\_17b\_16e & No Profile & 0.36 & 0.16 & 1.00 & 976.77 \\
phi4\_mini\_instruct & logical\_reasoner & 0.35 & 0.14 & 1.00 & 988.50 \\
mixtral\_8x7b & No Profile & 0.35 & 0.14 & 1.00 & 1000.19 \\
llama3.3\_70b & robust\_reasoner & 0.35 & 0.16 & 0.94 & 976.77 \\
llama3.3\_70b & No Profile & 0.35 & 0.14 & 1.00 & 988.50 \\
mixtral\_8x7b & robust\_reasoner & 0.33 & 0.12 & 1.00 & 1000.19 \\
llama3.3\_70b & logical\_reasoner & 0.33 & 0.14 & 0.94 & 1011.89 \\
phi4\_mini\_instruct & robust\_reasoner & 0.32 & 0.10 & 1.00 & 1000.19 \\
mixtral\_8x7b & logical\_reasoner & 0.32 & 0.12 & 0.94 & 988.50 \\
mixtral\_8x7b & deductive\_reasoner & 0.32 & 0.12 & 0.94 & 1000.19 \\
llama3.3\_70b & deductive\_reasoner & 0.32 & 0.12 & 0.94 & 1000.19 \\
llama3.2\_3b & logical\_reasoner & 0.29 & 0.06 & 1.00 & 976.77 \\
llama3.2\_3b & robust\_reasoner & 0.29 & 0.06 & 1.00 & 1000.19 \\
llama3.1\_8b & logical\_reasoner & 0.27 & 0.08 & 0.88 & 988.50 \\
llama3.1\_8b & robust\_reasoner & 0.27 & 0.12 & 0.80 & 953.15 \\
llama3.2\_3b & deductive\_reasoner & 0.27 & 0.04 & 1.00 & 1000.19 \\
llama3.2\_3b & No Profile & 0.27 & 0.06 & 0.94 & 1000.19 \\
        \hline
    \end{tabular}
    \caption{Experimental Results on the Calculus category of the OmniMATH dataset, Part 2 (continued from previous page)}
    \label{tab:cal2_Acc}
\end{table*}

\begin{table*}[htbp]
    \centering
    \begin{tabular}{@{}llcccc@{}}
        \hline
        \textbf{judge} & \textbf{profile} & \textbf{acc} & \textbf{SW\_acc} & \textbf{SR\_acc} & \textbf{Elo Score}\\
        \hline
gemma3\_12b & deductive\_reasoner & 0.76 & 0.79 & 0.53 & 1045.14 \\
gemma3\_12b & No Profile & 0.75 & 0.77 & 0.58 & 1055.4 \\
gemma3\_12b & logical\_reasoner & 0.74 & 0.76 & 0.58 & 1024.85 \\
gemma3\_12b & robust\_reasoner & 0.72 & 0.75 & 0.47 & 1076.19 \\
gemma3\_27b & deductive\_reasoner & 0.72 & 0.72 & 0.68 & 1045.14 \\
gemma3\_27b & logical\_reasoner & 0.68 & 0.69 & 0.58 & 1004.71 \\
gemma3\_4b & No Profile & 0.67 & 0.68 & 0.63 & 1004.71 \\
gemma3\_27b & robust\_reasoner & 0.67 & 0.66 & 0.78 & 994.66 \\
gemma3\_4b & robust\_reasoner & 0.66 & 0.7 & 0.37 & 994.66 \\
gemma3\_27b & No Profile & 0.64 & 0.64 & 0.67 & 1004.71 \\
gemma3\_4b & deductive\_reasoner & 0.62 & 0.62 & 0.58 & 1024.85 \\
gemma3\_4b & logical\_reasoner & 0.59 & 0.6 & 0.53 & 984.6 \\
qwen3\_4b & logical\_reasoner & 0.57 & 0.56 & 0.71 & 974.51 \\
qwen2.5\_72b & No Profile & 0.57 & 0.54 & 0.79 & 1045.14 \\
qwen3\_4b & deductive\_reasoner & 0.54 & 0.53 & 0.71 & 954.19 \\
qwen3\_4b & No Profile & 0.54 & 0.54 & 0.79 & 974.51 \\
qwen2.5\_72b & deductive\_reasoner & 0.49 & 0.44 & 0.84 & 974.51 \\
qwen3\_4b & robust\_reasoner & 0.48 & 0.46 & 0.69 & 943.93 \\
qwen3\_14b & deductive\_reasoner & 0.47 & 0.44 & 0.82 & 933.57 \\
qwen3\_14b & No Profile & 0.47 & 0.44 & 0.8 & 943.93 \\
qwen2.5\_72b & robust\_reasoner & 0.47 & 0.41 & 0.89 & 984.6 \\
qwen3\_8b & No Profile & 0.46 & 0.44 & 0.75 & 923.1 \\
qwen3\_14b & logical\_reasoner & 0.42 & 0.39 & 0.87 & 923.1 \\
qwen2.5\_72b & logical\_reasoner & 0.42 & 0.35 & 0.95 & 1024.85 \\
qwen3\_8b & logical\_reasoner & 0.42 & 0.38 & 0.87 & 890.77 \\
qwen3\_8b & deductive\_reasoner & 0.42 & 0.39 & 0.75 & 933.57 \\
qwen3\_14b & robust\_reasoner & 0.41 & 0.39 & 0.83 & 964.38 \\
llama3.3\_70b & robust\_reasoner & 0.4 & 0.32 & 0.95 & 1014.77 \\
phi4 & deductive\_reasoner & 0.38 & 0.31 & 0.89 & 1024.85 \\
phi4 & No Profile & 0.38 & 0.31 & 0.84 & 1024.85 \\
qwen3\_1.7b & deductive\_reasoner & 0.38 & 0.34 & 0.75 & 943.93 \\
phi4 & logical\_reasoner & 0.37 & 0.29 & 0.95 & 984.6 \\
qwen3\_1.7b & logical\_reasoner & 0.37 & 0.33 & 0.76 & 964.38 \\
llama3.1\_8b & No Profile & 0.37 & 0.31 & 0.83 & 974.51 \\
phi4 & robust\_reasoner & 0.36 & 0.3 & 0.84 & 1024.85 \\
qwen3\_8b & robust\_reasoner & 0.36 & 0.34 & 0.77 & 933.57 \\
qwen3\_0.6b & No Profile & 0.35 & 0.3 & 0.93 & 1045.14 \\
qwen3\_1.7b & robust\_reasoner & 0.35 & 0.33 & 0.62 & 912.49 \\
llama3.3\_70b & logical\_reasoner & 0.35 & 0.27 & 0.95 & 1004.71 \\
llama3.3\_70b & deductive\_reasoner & 0.35 & 0.29 & 0.84 & 994.66 \\
qwen3\_0.6b & robust\_reasoner & 0.35 & 0.29 & 0.88 & 1045.14 \\
llama3.3\_70b & No Profile & 0.35 & 0.28 & 0.84 & 1024.85 \\
qwen3\_0.6b & deductive\_reasoner & 0.34 & 0.29 & 0.82 & 1024.85 \\
llama\_4\_scout\_17b\_16e & logical\_reasoner & 0.34 & 0.25 & 1.0 & 1004.71 \\
llama3.1\_8b & deductive\_reasoner & 0.33 & 0.26 & 0.89 & 1045.14 \\
mixtral\_8x22b & logical\_reasoner & 0.32 & 0.24 & 0.89 & 1014.77 \\
        \hline
    \end{tabular}
    \caption{Experimental Results on the Discrete Math category of the OmniMATH dataset, Part 1}
    \label{tab:dis1_Acc}
\end{table*}

\begin{table*}[htbp]
    \centering
    \begin{tabular}{@{}llcccc@{}}
        \hline
        \textbf{judge} & \textbf{profile} & \textbf{acc} & \textbf{SW\_acc} & \textbf{SR\_acc} & \textbf{Elo Score}\\
        \hline
mixtral\_8x22b & deductive\_reasoner & 0.31 & 0.23 & 0.89 & 1024.85 \\
mixtral\_8x22b & robust\_reasoner & 0.31 & 0.23 & 0.89 & 1024.85 \\
llama3.1\_8b & logical\_reasoner & 0.31 & 0.23 & 0.89 & 994.66 \\
phi4\_mini\_instruct & deductive\_reasoner & 0.3 & 0.21 & 1.0 & 1024.85 \\
mixtral\_8x22b & No Profile & 0.3 & 0.22 & 0.89 & 1024.85 \\
qwen3\_0.6b & logical\_reasoner & 0.3 & 0.22 & 0.89 & 1024.85 \\
qwen3\_1.7b & No Profile & 0.3 & 0.28 & 0.6 & 943.93 \\
llama\_4\_scout\_17b\_16e & robust\_reasoner & 0.3 & 0.21 & 1.0 & 1014.77 \\
phi4\_mini\_instruct & logical\_reasoner & 0.3 & 0.2 & 1.0 & 1024.85 \\
mixtral\_8x7b & No Profile & 0.29 & 0.2 & 0.95 & 1034.97 \\
phi4\_mini\_instruct & No Profile & 0.28 & 0.19 & 1.0 & 1014.77 \\
llama\_4\_scout\_17b\_16e & deductive\_reasoner & 0.28 & 0.19 & 0.95 & 1014.77 \\
llama\_4\_scout\_17b\_16e & No Profile & 0.28 & 0.19 & 1.0 & 1014.77 \\
mixtral\_8x7b & logical\_reasoner & 0.27 & 0.18 & 0.95 & 1034.97 \\
llama3.1\_8b & robust\_reasoner & 0.26 & 0.18 & 0.84 & 974.51 \\
phi4\_mini\_instruct & robust\_reasoner & 0.25 & 0.15 & 1.0 & 1034.97 \\
mixtral\_8x7b & deductive\_reasoner & 0.23 & 0.13 & 1.0 & 1024.85 \\
mixtral\_8x7b & robust\_reasoner & 0.23 & 0.13 & 0.95 & 1024.85 \\
llama3.2\_3b & deductive\_reasoner & 0.19 & 0.08 & 1.0 & 1014.77 \\
llama3.2\_3b & No Profile & 0.19 & 0.08 & 1.0 & 1024.85 \\
llama3.2\_3b & robust\_reasoner & 0.18 & 0.07 & 0.95 & 1014.77 \\
llama3.2\_3b & logical\_reasoner & 0.16 & 0.05 & 1.0 & 1014.77 \\
        \hline
    \end{tabular}
    \caption{Experimental Results on the Discrete Math category of the OmniMATH dataset, Part 2 (continued from previous page)}
    \label{tab:dis2_Acc}
\end{table*}

\begin{table*}[htbp]
    \centering
    \begin{tabular}{@{}llcccc@{}}
        \hline
        \textbf{judge} & \textbf{profile} & \textbf{acc} & \textbf{SW\_acc} & \textbf{SR\_acc} & \textbf{Elo Score}\\
        \hline
gemma3\_12b & logical\_reasoner & 0.72 & 0.76 & 0.59 & 1018.2 \\
gemma3\_12b & robust\_reasoner & 0.72 & 0.78 & 0.54 & 1018.2 \\
gemma3\_27b & deductive\_reasoner & 0.72 & 0.73 & 0.68 & 1038.34 \\
gemma3\_4b & robust\_reasoner & 0.71 & 0.76 & 0.51 & 1058.73 \\
gemma3\_12b & No Profile & 0.7 & 0.75 & 0.54 & 1023.22 \\
gemma3\_4b & logical\_reasoner & 0.69 & 0.74 & 0.54 & 1048.5 \\
gemma3\_12b & deductive\_reasoner & 0.67 & 0.71 & 0.54 & 1048.5 \\
gemma3\_4b & No Profile & 0.66 & 0.7 & 0.54 & 1058.73 \\
gemma3\_4b & deductive\_reasoner & 0.65 & 0.66 & 0.62 & 1074.28 \\
gemma3\_27b & robust\_reasoner & 0.65 & 0.66 & 0.64 & 993.18 \\
gemma3\_27b & logical\_reasoner & 0.65 & 0.66 & 0.59 & 1043.41 \\
qwen3\_4b & logical\_reasoner & 0.64 & 0.65 & 0.72 & 1003.18 \\
gemma3\_27b & No Profile & 0.63 & 0.64 & 0.59 & 1013.19 \\
qwen3\_4b & deductive\_reasoner & 0.62 & 0.65 & 0.59 & 993.18 \\
qwen3\_4b & No Profile & 0.6 & 0.61 & 0.64 & 993.18 \\
qwen3\_4b & robust\_reasoner & 0.57 & 0.59 & 0.66 & 998.18 \\
qwen3\_1.7b & deductive\_reasoner & 0.56 & 0.55 & 0.69 & 957.98 \\
qwen2.5\_72b & No Profile & 0.53 & 0.48 & 0.7 & 1038.34 \\
qwen3\_14b & No Profile & 0.51 & 0.51 & 0.68 & 952.89 \\
qwen3\_1.7b & logical\_reasoner & 0.51 & 0.5 & 0.69 & 942.67 \\
qwen3\_14b & deductive\_reasoner & 0.51 & 0.5 & 0.68 & 968.09 \\
qwen3\_0.6b & robust\_reasoner & 0.49 & 0.39 & 0.83 & 942.67 \\
phi4 & logical\_reasoner & 0.49 & 0.34 & 0.95 & 983.17 \\
qwen3\_1.7b & No Profile & 0.49 & 0.45 & 0.74 & 957.98 \\
qwen3\_8b & deductive\_reasoner & 0.49 & 0.47 & 0.71 & 983.17 \\
qwen3\_8b & No Profile & 0.49 & 0.46 & 0.72 & 973.13 \\
qwen3\_0.6b & No Profile & 0.47 & 0.39 & 0.79 & 1013.19 \\
qwen3\_1.7b & robust\_reasoner & 0.47 & 0.44 & 0.72 & 947.79 \\
qwen3\_8b & logical\_reasoner & 0.47 & 0.45 & 0.76 & 963.04 \\
qwen2.5\_72b & deductive\_reasoner & 0.47 & 0.38 & 0.76 & 1023.22 \\
qwen3\_14b & logical\_reasoner & 0.46 & 0.45 & 0.72 & 921.9 \\
qwen2.5\_72b & robust\_reasoner & 0.46 & 0.36 & 0.78 & 1023.22 \\
phi4 & No Profile & 0.46 & 0.32 & 0.89 & 998.18 \\
qwen3\_14b & robust\_reasoner & 0.46 & 0.41 & 0.76 & 921.9 \\
phi4 & robust\_reasoner & 0.45 & 0.32 & 0.86 & 988.18 \\
qwen3\_0.6b & logical\_reasoner & 0.45 & 0.33 & 0.84 & 998.18 \\
qwen3\_0.6b & deductive\_reasoner & 0.45 & 0.34 & 0.83 & 998.18 \\
phi4 & deductive\_reasoner & 0.43 & 0.28 & 0.92 & 993.18 \\
qwen3\_8b & robust\_reasoner & 0.43 & 0.41 & 0.69 & 957.98 \\
qwen2.5\_72b & logical\_reasoner & 0.43 & 0.29 & 0.86 & 998.18 \\
llama\_4\_scout\_17b\_16e & logical\_reasoner & 0.4 & 0.26 & 0.86 & 998.18 \\
llama\_4\_scout\_17b\_16e & No Profile & 0.4 & 0.24 & 0.89 & 1013.19 \\
llama\_4\_scout\_17b\_16e & robust\_reasoner & 0.38 & 0.23 & 0.89 & 988.18 \\
llama3.3\_70b & robust\_reasoner & 0.38 & 0.24 & 0.84 & 1023.22 \\
llama3.1\_8b & No Profile & 0.38 & 0.25 & 0.81 & 993.18 \\
llama\_4\_scout\_17b\_16e & deductive\_reasoner & 0.38 & 0.24 & 0.84 & 1008.19 \\
        \hline
    \end{tabular}
    \caption{Experimental Results on the Number Theory category of the OmniMATH dataset, Part 1}
    \label{tab:nt1_Acc}
\end{table*}

\begin{table*}[htbp]
    \centering
    \begin{tabular}{@{}llcccc@{}}
        \hline
        \textbf{judge} & \textbf{profile} & \textbf{acc} & \textbf{SW\_acc} & \textbf{SR\_acc} & \textbf{Elo Score}\\
        \hline
llama3.3\_70b & logical\_reasoner & 0.38 & 0.22 & 0.89 & 998.18 \\
phi4\_mini\_instruct & deductive\_reasoner & 0.37 & 0.21 & 0.89 & 1008.19 \\
llama3.1\_8b & deductive\_reasoner & 0.37 & 0.27 & 0.72 & 1003.18 \\
llama3.3\_70b & No Profile & 0.37 & 0.24 & 0.78 & 1003.18 \\
mixtral\_8x22b & logical\_reasoner & 0.35 & 0.18 & 0.92 & 1003.18 \\
llama3.1\_8b & robust\_reasoner & 0.35 & 0.19 & 0.91 & 988.18 \\
phi4\_mini\_instruct & No Profile & 0.35 & 0.15 & 0.97 & 988.18 \\
mixtral\_8x7b & logical\_reasoner & 0.35 & 0.16 & 0.95 & 1023.22 \\
mixtral\_8x22b & robust\_reasoner & 0.35 & 0.15 & 0.97 & 998.18 \\
llama3.3\_70b & deductive\_reasoner & 0.35 & 0.19 & 0.84 & 1008.19 \\
mixtral\_8x22b & No Profile & 0.34 & 0.15 & 0.95 & 1008.19 \\
mixtral\_8x22b & deductive\_reasoner & 0.33 & 0.14 & 0.95 & 998.18 \\
llama3.1\_8b & logical\_reasoner & 0.33 & 0.16 & 0.89 & 983.17 \\
phi4\_mini\_instruct & logical\_reasoner & 0.32 & 0.13 & 0.95 & 993.18 \\
mixtral\_8x7b & No Profile & 0.32 & 0.13 & 0.92 & 1013.19 \\
phi4\_mini\_instruct & robust\_reasoner & 0.3 & 0.11 & 0.92 & 1018.2 \\
mixtral\_8x7b & robust\_reasoner & 0.3 & 0.1 & 0.95 & 1028.25 \\
llama3.2\_3b & logical\_reasoner & 0.3 & 0.08 & 1.0 & 1013.19 \\
llama3.2\_3b & No Profile & 0.29 & 0.08 & 0.97 & 1003.18 \\
mixtral\_8x7b & deductive\_reasoner & 0.29 & 0.08 & 0.97 & 1023.22 \\
llama3.2\_3b & deductive\_reasoner & 0.29 & 0.07 & 1.0 & 1018.2 \\
llama3.2\_3b & robust\_reasoner & 0.27 & 0.05 & 0.97 & 983.17 \\
        \hline
    \end{tabular}
    \caption{Experimental Results on the Number Theory category of the OmniMATH dataset, Part 2 (continued from previous page)}
    \label{tab:nt2_Acc}
\end{table*}

\begin{table*}
    \centering
    \begin{tabular}{@{}lcccc@{}}
        \hline
        \textbf{judge} & \textbf{acc} & \textbf{SW\_acc} & \textbf{SR\_acc} & \textbf{Elo Score}\\
        \hline
gemma3\_4b & 0.31 & 0.18 & 0.87 & 1005.44 \\
mixtral\_8x7b & 0.30 & 0.13 & 1.00 & 1005.44 \\
llama3.2\_3b & 0.29 & 0.11 & 1.00 & 993.82 \\
qwen3\_0.6b & 0.26 & 0.10 & 0.93 & 1017.07 \\
gemma3\_12b & 0.26 & 0.10 & 0.93 & 1017.07 \\
qwen3\_14b & 0.25 & 0.06 & 1.00 & 982.18 \\
qwen2.5\_72b & 0.25 & 0.06 & 1.00 & 993.82 \\
qwen3\_4b & 0.23 & 0.05 & 1.00 & 1005.44 \\
qwen3\_8b & 0.23 & 0.05 & 1.00 & 1005.44 \\
gemma3\_27b & 0.23 & 0.05 & 1.00 & 982.18 \\
phi4 & 0.22 & 0.03 & 1.00 & - \\
qwen3\_1.7b & 0.22 & 0.05 & 0.93 & 1040.48 \\
llama3.1\_8b & 0.22 & 0.03 & 1.00 & 946.92 \\
llama\_4\_scout\_17b\_16e & 0.22 & 0.03 & 1.00 & 993.82 \\
mixtral\_8x22b & 0.21 & 0.02 & 1.00 & 1017.07 \\
llama3.3\_70b & 0.19 & 0.00 & 1.00 & 993.82 \\
        \hline
    \end{tabular}
    \caption{Experimental Results on the Biology Category of the GPQA dataset}
    \label{tab:Bio_Acc}
\end{table*}

\begin{table*}
    \centering
    \begin{tabular}{@{}lcccc@{}}
        \hline
        \textbf{judge} & \textbf{acc} & \textbf{SW\_acc} & \textbf{SR\_acc} & \textbf{Elo Score}\\
        \hline
gemma3\_4b & 0.303 & 0.217 & 0.750 & 1005.08 \\
mixtral\_8x7b & 0.162 & 0.000 & 1.000 & 1005.08 \\
llama3.2\_3b & 0.172 & 0.012 & 1.000 & 994.2 \\
qwen3\_0.6b & 0.162 & 0.048 & 0.750 & 1026.89 \\
gemma3\_12b & 0.384 & 0.277 & 0.938 & 994.2 \\
qwen3\_14b & 0.343 & 0.217 & 1.000 & 961.42 \\
qwen2.5\_72b & 0.232 & 0.096 & 0.938 & 1015.97 \\
qwen3\_4b & 0.313 & 0.181 & 1.000 & 1005.08 \\
qwen3\_8b & 0.303 & 0.169 & 1.000 & 994.2 \\
gemma3\_27b & 0.192 & 0.048 & 0.938 & 1015.97 \\
phi4 & 0.192 & 0.036 & 1.000 & - \\
qwen3\_1.7b & 0.232 & 0.108 & 0.875 & 983.32 \\
llama3.1\_8b & 0.141 & 0.012 & 0.867 & 994.2 \\
llama\_4\_scout\_17b\_16e & 0.162 & 0.000 & 1.000 & 1005.08 \\
mixtral\_8x22b & 0.182 & 0.024 & 1.000 & 994.2 \\
llama3.3\_70b & 0.162 & 0.000 & 1.000 & 1005.08 \\
        \hline
    \end{tabular}
    \caption{Experimental Results on the Chemistry Category of the GPQA dataset}
    \label{tab:Chem_Acc}
\end{table*}

\begin{table*}
    \centering
    \begin{tabular}{@{}lcccc@{}}
        \hline
        \textbf{judge} & \textbf{acc} & \textbf{SW\_acc} & \textbf{SR\_acc} & \textbf{Elo Score}\\
        \hline
qwen3\_0.6b & 0.43 & 0.11 & 0.97 & 981.44 \\
qwen3\_14b & 0.43 & 0.11 & 1.00 & 1013.92 \\
gemma3\_4b & 0.43 & 0.18 & 0.86 & 1004.64 \\
qwen3\_4b & 0.41 & 0.08 & 0.97 & 1000.00 \\
gemma3\_12b & 0.41 & 0.18 & 0.81 & 986.08 \\
gemma3\_27b & 0.41 & 0.08 & 0.97 & 986.08 \\
mixtral\_8x7b & 0.39 & 0.03 & 1.00 & 1009.28 \\
qwen3\_8b & 0.39 & 0.03 & 1.00 & 1009.28 \\
mixtral\_8x22b & 0.38 & 0.03 & 0.97 & 1013.92 \\
qwen3\_1.7b & 0.38 & 0.05 & 0.95 & 1000.00 \\
llama3.1\_8b & 0.38 & 0.02 & 1.00 & 990.72 \\
llama\_4\_scout\_17b\_16e & 0.38 & 0.02 & 1.00 & - \\
llama3.3\_70b & 0.38 & 0.02 & 1.00 & - \\
qwen2.5\_72b & 0.38 & 0.02 & 1.00 & 1004.64 \\
phi4 & 0.37 & 0.00 & 1.00 & - \\
llama3.2\_3b & 0.37 & 0.02 & 0.97 & 1000.00 \\
        \hline
    \end{tabular}
    \caption{Experimental Results on the Physics Category of the GPQA dataset}
    \label{tab:Phys_Acc}
\end{table*}

%% file: aaai2026.bib
@article{tang2024large,
  title={Large Language Models for Automated Literature Review: An Evaluation of Reference Generation, Abstract Writing, and Review Composition},
  author={Tang, Xuemei and Duan, Xufeng and Cai, Zhenguang G},
  journal={arXiv preprint arXiv:2412.13612},
  year={2024}
}

@article{tan2024judgebench,
  title={Judgebench: A benchmark for evaluating llm-based judges},
  author={Tan, Sijun and Zhuang, Siyuan and Montgomery, Kyle and Tang, William Y and Cuadron, Alejandro and Wang, Chenguang and Popa, Raluca Ada and Stoica, Ion},
  journal={arXiv preprint arXiv:2410.12784},
  year={2024}
}

@article{zheng2023judging,
  title={Judging llm-as-a-judge with mt-bench and chatbot arena},
  author={Zheng, Lianmin and Chiang, Wei-Lin and Sheng, Ying and Zhuang, Siyuan and Wu, Zhanghao and Zhuang, Yonghao and Lin, Zi and Li, Zhuohan and Li, Dacheng and Xing, Eric and others},
  journal={Advances in neural information processing systems},
  volume={36},
  pages={46595--46623},
  year={2023}
}

@article{wei2022chain,
  title={Chain-of-thought prompting elicits reasoning in large language models},
  author={Wei, Jason and Wang, Xuezhi and Schuurmans, Dale and Bosma, Maarten and Xia, Fei and Chi, Ed and Le, Quoc V and Zhou, Denny and others},
  journal={Advances in neural information processing systems},
  volume={35},
  pages={24824--24837},
  year={2022}
}

@article{wang2022self,
  title={Self-consistency improves chain of thought reasoning in language models},
  author={Wang, Xuezhi and Wei, Jason and Schuurmans, Dale and Le, Quoc and Chi, Ed and Narang, Sharan and Chowdhery, Aakanksha and Zhou, Denny},
  journal={arXiv preprint arXiv:2203.11171},
  year={2022}
}

@article{yao2023tree,
  title={Tree of thoughts: Deliberate problem solving with large language models},
  author={Yao, Shunyu and Yu, Dian and Zhao, Jeffrey and Shafran, Izhak and Griffiths, Tom and Cao, Yuan and Narasimhan, Karthik},
  journal={Advances in neural information processing systems},
  volume={36},
  pages={11809--11822},
  year={2023}
}

@inproceedings{besta2024graph,
  title={Graph of thoughts: Solving elaborate problems with large language models},
  author={Besta, Maciej and Blach, Nils and Kubicek, Ales and Gerstenberger, Robert and Podstawski, Michal and Gianinazzi, Lukas and Gajda, Joanna and Lehmann, Tomasz and Niewiadomski, Hubert and Nyczyk, Piotr and others},
  booktitle={Proceedings of the AAAI conference on artificial intelligence},
  volume={38},
  number={16},
  pages={17682--17690},
  year={2024}
}

@inproceedings{wu2024autogen,
  title={Autogen: Enabling next-gen LLM applications via multi-agent conversations},
  author={Wu, Qingyun and Bansal, Gagan and Zhang, Jieyu and Wu, Yiran and Li, Beibin and Zhu, Erkang and Jiang, Li and Zhang, Xiaoyun and Zhang, Shaokun and Liu, Jiale and others},
  booktitle={First Conference on Language Modeling},
  year={2024}
}

@article{chen2023autoagents,
  title={Autoagents: A framework for automatic agent generation},
  author={Chen, Guangyao and Dong, Siwei and Shu, Yu and Zhang, Ge and Sesay, Jaward and Karlsson, B{\"o}rje F and Fu, Jie and Shi, Yemin},
  journal={arXiv preprint arXiv:2309.17288},
  year={2023}
}

@article{zhugegptswarm,
  title={Language Agents as Optimizable Graphs},
  author={Mingchen Zhuge and Wenyi Wang and Louis Kirsch and Francesco Faccio and Dmitrii Khizbullin and J{\"u}rgen Schmidhuber},
  journal={ArXiv},
  year={2024},
  volume={abs/2402.16823},
  url={https://api.semanticscholar.org/CorpusID:268032156}
}

@article{dubey2024llama,
  title={The llama 3 herd of models},
  author={Dubey, Abhimanyu and Jauhri, Abhinav and Pandey, Abhinav and Kadian, Abhishek and Al-Dahle, Ahmad and Letman, Aiesha and Mathur, Akhil and Schelten, Alan and Yang, Amy and Fan, Angela and others},
  journal={arXiv e-prints},
  pages={arXiv--2407},
  year={2024}
}

@article{liu2024deepseek,
  title={Deepseek-v3 technical report},
  author={Liu, Aixin and Feng, Bei and Xue, Bing and Wang, Bingxuan and Wu, Bochao and Lu, Chengda and Zhao, Chenggang and Deng, Chengqi and Zhang, Chenyu and Ruan, Chong and others},
  journal={arXiv preprint arXiv:2412.19437},
  year={2024}
}

@article{achiam2023gpt,
  title={Gpt-4 technical report},
  author={Achiam, Josh and Adler, Steven and Agarwal, Sandhini and Ahmad, Lama and Akkaya, Ilge and Aleman, Florencia Leoni and Almeida, Diogo and Altenschmidt, Janko and Altman, Sam and Anadkat, Shyamal and others},
  journal={arXiv preprint arXiv:2303.08774},
  year={2023}
}

@article{yang2025qwen3,
  title={Qwen3 technical report},
  author={Yang, An and Li, Anfeng and Yang, Baosong and Zhang, Beichen and Hui, Binyuan and Zheng, Bo and Yu, Bowen and Gao, Chang and Huang, Chengen and Lv, Chenxu and others},
  journal={arXiv preprint arXiv:2505.09388},
  year={2025}
}

@article{abdin2024phi,
  title={Phi-4 technical report},
  author={Abdin, Marah and Aneja, Jyoti and Behl, Harkirat and Bubeck, S{\'e}bastien and Eldan, Ronen and Gunasekar, Suriya and Harrison, Michael and Hewett, Russell J and Javaheripi, Mojan and Kauffmann, Piero and others},
  journal={arXiv preprint arXiv:2412.08905},
  year={2024}
}

@article{team2024gemma,
  title={Gemma: Open models based on gemini research and technology},
  author={Team, Gemma and Mesnard, Thomas and Hardin, Cassidy and Dadashi, Robert and Bhupatiraju, Surya and Pathak, Shreya and Sifre, Laurent and Rivi{\`e}re, Morgane and Kale, Mihir Sanjay and Love, Juliette and others},
  journal={arXiv preprint arXiv:2403.08295},
  year={2024}
}

@article{guo2024large,
  title={Large language model based multi-agents: A survey of progress and challenges},
  author={Guo, Taicheng and Chen, Xiuying and Wang, Yaqi and Chang, Ruidi and Pei, Shichao and Chawla, Nitesh V and Wiest, Olaf and Zhang, Xiangliang},
  journal={arXiv preprint arXiv:2402.01680},
  year={2024}
}

@article{lightman2023lets,
      title={Let's Verify Step by Step}, 
      author={Lightman, Hunter and Kosaraju, Vineet and Burda, Yura and Edwards, Harri and Baker, Bowen and Lee, Teddy and Leike, Jan and Schulman, John and Sutskever, Ilya and Cobbe, Karl},
      journal={arXiv preprint arXiv:2305.20050},
      year={2023}
}

@article{cobbe2021training,
  title={Training verifiers to solve math word problems},
  author={Cobbe, Karl and Kosaraju, Vineet and Bavarian, Mohammad and Chen, Mark and Jun, Heewoo and Kaiser, Lukasz and Plappert, Matthias and Tworek, Jerry and Hilton, Jacob and Nakano, Reiichiro and others},
  journal={arXiv preprint arXiv:2110.14168},
  year={2021}
}

@article{li2024gsm,
  title={Gsm-plus: A comprehensive benchmark for evaluating the robustness of llms as mathematical problem solvers},
  author={Li, Qintong and Cui, Leyang and Zhao, Xueliang and Kong, Lingpeng and Bi, Wei},
  journal={arXiv preprint arXiv:2402.19255},
  year={2024}
}

@article{hendrycks2021measuring,
  title={Measuring mathematical problem solving with the math dataset},
  author={Hendrycks, Dan and Burns, Collin and Kadavath, Saurav and Arora, Akul and Basart, Steven and Tang, Eric and Song, Dawn and Steinhardt, Jacob},
  journal={arXiv preprint arXiv:2103.03874},
  year={2021}
}

@article{gao2024omni,
  title={Omni-math: A universal olympiad level mathematic benchmark for large language models},
  author={Gao, Bofei and Song, Feifan and Yang, Zhe and Cai, Zefan and Miao, Yibo and Dong, Qingxiu and Li, Lei and Ma, Chenghao and Chen, Liang and Xu, Runxin and others},
  journal={arXiv preprint arXiv:2410.07985},
  year={2024}
}

@article{clark2018think,
  title={Think you have solved question answering? try arc, the ai2 reasoning challenge},
  author={Clark, Peter and Cowhey, Isaac and Etzioni, Oren and Khot, Tushar and Sabharwal, Ashish and Schoenick, Carissa and Tafjord, Oyvind},
  journal={arXiv preprint arXiv:1803.05457},
  year={2018}
}

@inproceedings{rein2024gpqa,
  title={Gpqa: A graduate-level google-proof q\&a benchmark},
  author={Rein, David and Hou, Betty Li and Stickland, Asa Cooper and Petty, Jackson and Pang, Richard Yuanzhe and Dirani, Julien and Michael, Julian and Bowman, Samuel R},
  booktitle={First Conference on Language Modeling},
  year={2024}
}

@misc{jiang2023mistral,
    title={Mistral 7B},
    author={Albert Q. Jiang and Alexandre Sablayrolles and Arthur Mensch and Chris Bamford and Devendra Singh Chaplot and Diego de las Casas and Florian Bressand and Gianna Lengyel and Guillaume Lample and Lucile Saulnier and Lélio Renard Lavaud and Marie-Anne Lachaux and Pierre Stock and Teven Le Scao and Thibaut Lavril and Thomas Wang and Timothée Lacroix and William El Sayed},
    year={2023},
    eprint={2310.06825},
    archivePrefix={arXiv},
    primaryClass={cs.CL}
}

@misc{myself,
      title={STOC-TOT: Stochastic Tree-of-Thought with Constrained Decoding for Complex Reasoning in Multi-Hop Question Answering}, 
      author={Zhenyu Bi and Daniel Hajialigol and Zhongkai Sun and Jie Hao and Xuan Wang},
      year={2024},
      eprint={2407.03687},
      archivePrefix={arXiv},
      primaryClass={cs.CL},
      url={https://arxiv.org/abs/2407.03687}, 
}

@misc{aot,
      title={Algorithm of Thoughts: Enhancing Exploration of Ideas in Large Language Models}, 
      author={Bilgehan Sel and Ahmad Al-Tawaha and Vanshaj Khattar and Ruoxi Jia and Ming Jin},
      year={2024},
      eprint={2308.10379},
      archivePrefix={arXiv},
      primaryClass={cs.CL},
      url={https://arxiv.org/abs/2308.10379}, 
}

@misc{pot,
      title={Program of Thoughts Prompting: Disentangling Computation from Reasoning for Numerical Reasoning Tasks}, 
      author={Wenhu Chen and Xueguang Ma and Xinyi Wang and William W. Cohen},
      year={2023},
      eprint={2211.12588},
      archivePrefix={arXiv},
      primaryClass={cs.CL},
      url={https://arxiv.org/abs/2211.12588}, 
}

@inproceedings{menglu,
    title = "{T}riage{A}gent: Towards Better Multi-Agents Collaborations for Large Language Model-Based Clinical Triage",
    author = "Lu, Meng  and
      Ho, Brandon  and
      Ren, Dennis  and
      Wang, Xuan",
    booktitle = "Findings of the Association for Computational Linguistics: EMNLP 2024",
    month = nov,
    year = "2024",
    address = "Miami, Florida, USA",
    publisher = "Association for Computational Linguistics",
    url = "https://aclanthology.org/2024.findings-emnlp.329/",
    doi = "10.18653/v1/2024.findings-emnlp.329",
    pages = "5747--5764",
}

@article{Wang2023ASO,
  title={A Survey on Large Language Model based Autonomous Agents},
  author={Lei Wang and Chengbang Ma and Xueyang Feng and Zeyu Zhang and Hao-ran Yang and Jingsen Zhang and Zhi-Yang Chen and Jiakai Tang and Xu Chen and Yankai Lin and Wayne Xin Zhao and Zhewei Wei and Ji-rong Wen},
  journal={ArXiv},
  year={2023},
  volume={abs/2308.11432},
  url={https://api.semanticscholar.org/CorpusID:261064713}
}

@inproceedings{Wang2024RethinkingTB,
  title={Rethinking the Bounds of LLM Reasoning: Are Multi-Agent Discussions the Key?},
  author={Qineng Wang and Zihao Wang and Ying Su and Hanghang Tong and Yangqiu Song},
  booktitle={Annual Meeting of the Association for Computational Linguistics},
  year={2024},
  url={https://api.semanticscholar.org/CorpusID:268041461}
}

@article{Pezeshkpour2024ReasoningCI,
  title={Reasoning Capacity in Multi-Agent Systems: Limitations, Challenges and Human-Centered Solutions},
  author={Pouya Pezeshkpour and Eser Kandogan and Nikita Bhutani and Sajjadur Rahman and Tom Mitchell and Estevam R. Hruschka},
  journal={ArXiv},
  year={2024},
  volume={abs/2402.01108},
  url={https://api.semanticscholar.org/CorpusID:267406749}
}

@inproceedings{Wu2023AutoGenEN,
  title={AutoGen: Enabling Next-Gen LLM Applications via Multi-Agent Conversation},
  author={Qingyun Wu and Gagan Bansal and Jieyu Zhang and Yiran Wu and Beibin Li and Erkang Zhu and Li Jiang and Xiaoyun Zhang and Shaokun Zhang and Jiale Liu and Ahmed Hassan Awadallah and Ryen W. White and Doug Burger and Chi Wang},
  year={2023},
  url={https://api.semanticscholar.org/CorpusID:263611068}
}

@misc{liang2024encouragingdivergentthinkinglarge,
      title={Encouraging Divergent Thinking in Large Language Models through Multi-Agent Debate}, 
      author={Tian Liang and Zhiwei He and Wenxiang Jiao and Xing Wang and Yan Wang and Rui Wang and Yujiu Yang and Shuming Shi and Zhaopeng Tu},
      year={2024},
      eprint={2305.19118},
      archivePrefix={arXiv},
      primaryClass={cs.CL},
      url={https://arxiv.org/abs/2305.19118}, 
}

@article{Chan2023ChatEvalTB,
  title={ChatEval: Towards Better LLM-based Evaluators through Multi-Agent Debate},
  author={Chi-Min Chan and Weize Chen and Yusheng Su and Jianxuan Yu and Wei Xue and Shan Zhang and Jie Fu and Zhiyuan Liu},
  journal={ArXiv},
  year={2023},
  volume={abs/2308.07201},
  url={https://api.semanticscholar.org/CorpusID:260887105}
}

@article{Chen2023ReConcileRC,
  title={ReConcile: Round-Table Conference Improves Reasoning via Consensus among Diverse LLMs},
  author={Justin Chih-Yao Chen and Swarnadeep Saha and Mohit Bansal},
  journal={ArXiv},
  year={2023},
  volume={abs/2309.13007},
  url={https://api.semanticscholar.org/CorpusID:262217323}
}

@article{GPT35,
  title={Language Models are Few-Shot Learners},
  author={Tom B. Brown and Benjamin Mann and Nick Ryder and Melanie Subbiah and Jared Kaplan and Prafulla Dhariwal and Arvind Neelakantan and Pranav Shyam and Girish Sastry and Amanda Askell and Sandhini Agarwal and Ariel Herbert-Voss and Gretchen Krueger and Tom Henighan and Rewon Child and Aditya Ramesh and Daniel M. Ziegler and Jeff Wu and Clemens Winter and Christopher Hesse and Mark Chen and Eric Sigler and Ma-teusz Litwin and Scott Gray and Benjamin Chess and Jack Clark and Christopher Berner and Sam McCandlish and Alec Radford and Ilya Sutskever and Dario Amodei},
  journal={ArXiv},
  year={2020},
  volume={abs/2005.14165},
  url={https://api.semanticscholar.org/CorpusID:218971783}
}

@inproceedings{GPT4,
  title={GPT-4 Technical Report},
  author={OpenAI},
  year={2023},
  url={https://api.semanticscholar.org/CorpusID:257532815}
}

@article{gunasekar2023textbooks,
  title={Textbooks are all you need},
  author={Gunasekar, Suriya and Zhang, Yi and Aneja, Jyoti and Mendes, Caio C{\'e}sar Teodoro and Del Giorno, Allie and Gopi, Sivakanth and Javaheripi, Mojan and Kauffmann, Piero and de Rosa, Gustavo and Saarikivi, Olli and others},
  journal={arXiv preprint arXiv:2306.11644},
  year={2023}
}

@article{bai2023qwen,
  title={Qwen technical report},
  author={Bai, Jinze and Bai, Shuai and Chu, Yunfei and Cui, Zeyu and Dang, Kai and Deng, Xiaodong and Fan, Yang and Ge, Wenbin and Han, Yu and Huang, Fei and others},
  journal={arXiv preprint arXiv:2309.16609},
  year={2023}
}

@article{tang2024rethinking,
  title={Rethinking the role of scale for in-context learning: An interpretability-based case study at 66 billion scale},
  author={Hritik Bansal and Karthik Gopalakrishnan and Saket Dingliwal and Sravan Bodapati and Katrin Kirchhoff and Dan Roth},
  journal={arXiv preprint arXiv:2212.09095},
  year={2024}
}

@article{liu2024makes,
  title={What makes good data for alignment? A comprehensive study of automatic data selection in instruction tuning},
  author={Liu, Wei and Zeng, Weihao and He, Keqing and Jiang, Yong and He, Junxian},
  journal={arXiv preprint arXiv:2312.15685},
  year={2024}
}

@article{shao2024deepseekmath,
  title={DeepSeekMath: Pushing the limits of mathematical reasoning in open language models},
  author={Shao, Zhihong and Wang, Peiyi and Zhu, Qihao and Xu, Runxin and Song, Junxiao and Zhang, Mingchuan and Li, YK and Wu, Y and Guo, Daya},
  journal={arXiv preprint arXiv:2402.03300},
  year={2024}
}

@inproceedings{srivastava-etal-2025-thinkslm,
    title = "{T}hink{SLM}: Towards Reasoning in Small Language Models",
    author = "Srivastava, Gaurav  and
      Cao, Shuxiang  and
      Wang, Xuan",
    booktitle = "Proceedings of the 2025 Conference on Empirical Methods in Natural Language Processing",
    month = nov,
    year = "2025",
    address = "Suzhou, China",
    publisher = "Association for Computational Linguistics",
    url = "https://aclanthology.org/2025.emnlp-main.1659/",
    doi = "10.18653/v1/2025.emnlp-main.1659",
    pages = "32600--32650",
}

@misc{bi2025optagentoptimizingmultiagentllm,
      title={OPTAGENT: Optimizing Multi-Agent LLM Interactions Through Verbal Reinforcement Learning for Enhanced Reasoning}, 
      author={Zhenyu Bi and Meng Lu and Yang Li and Swastik Roy and Weijie Guan and Morteza Ziyadi and Xuan Wang},
      year={2025},
      eprint={2510.18032},
      archivePrefix={arXiv},
      primaryClass={cs.AI},
      url={https://arxiv.org/abs/2510.18032}, 
}

@inproceedings{srivastava-etal-2025-debate,
    title = "{DEBATE}, {TRAIN}, {EVOLVE}: {S}elf{-}{E}volution of Language Model Reasoning",
    author = "Srivastava, Gaurav  and
      Bi, Zhenyu  and
      Lu, Meng  and
      Wang, Xuan",
    booktitle = "Proceedings of the 2025 Conference on Empirical Methods in Natural Language Processing",
    month = nov,
    year = "2025",
    address = "Suzhou, China",
    publisher = "Association for Computational Linguistics",
    url = "https://aclanthology.org/2025.emnlp-main.1666/",
    doi = "10.18653/v1/2025.emnlp-main.1666",
    pages = "32752--32798",
}

@misc{srivastava2025beyondbenchbenchmarkfreeevaluationreasoning,
      title={BeyondBench: Benchmark-Free Evaluation of Reasoning in Language Models}, 
      author={Gaurav Srivastava and Aafiya Hussain and Zhenyu Bi and Swastik Roy and Priya Pitre and Meng Lu and Morteza Ziyadi and Xuan Wang},
      year={2025},
      eprint={2509.24210},
      archivePrefix={arXiv},
      primaryClass={cs.CL},
      url={https://arxiv.org/abs/2509.24210}, 
}

@misc{srivastava2025llmsoverthinkbasicmath,
      title={Do LLMs Overthink Basic Math Reasoning? Benchmarking the Accuracy-Efficiency Tradeoff in Language Models}, 
      author={Gaurav Srivastava and Aafiya Hussain and Sriram Srinivasan and Xuan Wang},
      year={2025},
      eprint={2507.04023},
      archivePrefix={arXiv},
      primaryClass={cs.CL},
      url={https://arxiv.org/abs/2507.04023}, 
}
